\documentclass{article}

 \usepackage[dblblindworkshop, final]{neurips_2025}

\workshoptitle{Bridging Language, Agent, and World Models for Reasoning and Planning (LAW)}
\usepackage{multicol}
\usepackage{enumitem}
\usepackage[utf8]{inputenc} 
\usepackage[T1]{fontenc}    
\usepackage{hyperref}       
\usepackage{url}  
\usepackage{multicol}
\usepackage{booktabs}       
\usepackage{amsfonts}       
\usepackage{nicefrac}       
\usepackage{microtype}      
\usepackage{xcolor}         

\usepackage{adjustbox}
\usepackage{verbatim}
\usepackage{listings}
\usepackage{natbib}
\usepackage{graphicx}
\usepackage{wrapfig}
\usepackage{tcolorbox}
\usepackage{amsmath}
\usepackage{booktabs}
\usepackage{array}
\tcbuselibrary{listings, skins, breakable}
\usepackage{subcaption}
\usepackage{float}
\usepackage{capt-of}   
\usepackage[normalem]{ulem}
\usepackage{xcolor} 
\usepackage{hyperref}
\usepackage[normalem]{ulem}

\definecolor{mycitecolor}{HTML}{6da7c9}
\definecolor{myyellow}{HTML}{FDD33A}
\definecolor{myplan}{HTML}{FDD33A}
\definecolor{morange}{HTML}{e48c3f}
\definecolor{mgreen}{HTML}{47ba78}
\definecolor{mred}{HTML}{eb6759}
\definecolor{mpurple}{HTML}{c9c0e6}
\hypersetup{
  colorlinks=true,     
  linkcolor=black,     
  citecolor=mycitecolor, 
  urlcolor=black       
}
\newcommand{\jes}[1]{{\color{black} #1}}

\title{DR. WELL: Dynamic Reasoning and Learning with Symbolic World Model for Embodied LLM-Based Multi-Agent Collaboration}

\author{
  Narjes Nourzad\thanks{Equal contribution.}\,\,\,\thanks{Work done during an internship at Carnegie Mellon University (CMU).} \\
  University of Southern California \\
  \texttt{nourzad@usc.edu} \\
  \And
  Hanqing Yang\footnotemark[1] \\
  Carnegie Mellon University \\
  \texttt{hanqing3@andrew.cmu.edu} \\
  \And
  Shiyu Chen \\
  Carnegie Mellon University \\
  \texttt{shiyuc@andrew.cmu.edu} \\
  \And
  Carlee Joe-Wong \\
  Carnegie Mellon University \\
  \texttt{cjoewong@andrew.cmu.edu} \\
}

\begin{document}

\maketitle

\begin{abstract}
Cooperative multi-agent planning requires agents to make joint decisions with partial information and limited communication. Coordination at the trajectory level often fails, as small deviations in timing or movement cascade into conflicts. Symbolic planning mitigates this challenge by raising the level of abstraction and providing a minimal vocabulary of actions that enable synchronization and collective progress.
We present DR. WELL, a decentralized neurosymbolic framework for cooperative multi-agent planning. Cooperation unfolds through a two-phase negotiation protocol: agents first propose candidate roles with reasoning and then commit to a joint allocation under consensus and environment constraints. After commitment, each agent independently generates and executes a symbolic plan for its role without revealing detailed trajectories. Plans are grounded in execution outcomes via a shared world model that encodes the current state and is updated as agents act.
By reasoning over symbolic plans rather than raw trajectories, DR. WELL avoids brittle step-level alignment and enables higher-level operations that are reusable, synchronizable, and interpretable. Experiments on cooperative block-push tasks show that agents adapt across episodes, with the dynamic world model capturing reusable patterns and improving task completion rates and efficiency. Experiments on cooperative block-push tasks show that our dynamic world model improves task completion and efficiency through negotiation and self-refinement, trading a time overhead for evolving, more efficient collaboration strategies. The project is open-sourced at: \url{https://narjesno.github.io/DR.WELL/}.

\end{abstract}

\section{Introduction}
Cooperation lies at the heart of many multi-agent problems. 
Whether autonomous vehicles are coordinating at intersections~\citep{wu2019dcl} or distributed sensor networks are monitoring large environments~\citep{mishra2024multi}, success depends on dividing tasks and aligning agents' intentions toward shared objectives.
Despite advances in both single- and multi-agent reinforcement learning (MARL), achieving generalizable cooperative behavior beyond narrow training scenarios remains difficult to model effectively~\citep{jin2025comprehensive, ning2024survey, zhuge2024gptswarm}.
Recent work has introduced large language models (LLMs) into multi-agent reinforcement learning to improve flexibility and generalization~\citep{yang2023foundation, ma2025agentic, chen2025self, yang2025llm}.  
For cooperation, LLMs can translate group objectives into role-specific subgoals, describe coordination requirements, and adapt allocations as conditions change~\citep{auranourzad}.  
In embodied settings with decentralized execution, however, agents face additional challenges since they must coordinate under partial observability, limited bandwidth, and asynchronous timing without centralized control.  
In these conditions, directly binding agent policies to raw LLM outputs has proven fragile, since behaviors depend heavily on prompt phrasing and often fail to generalize when the number of agents or environment conditions change~\citep{shah2025learning}.
LLMs are therefore better used in combination with structured representations, which can provide stability and consistency in cooperative planning.  
One natural choice is to equip agents with a shared vocabulary of \textit{concepts} that complements LLM guidance.  
This design reduces reliance on fragile prompting while allowing coordination to emerge through composing and exchanging these symbolic structures, rather than through a single joint policy, thereby improving generalization~\citep{zhou2024symbolic, inala2020neurosymbolic}.

\begin{wrapfigure}{r}{0.5\textwidth}
  \centering
  \includegraphics[width=\linewidth]{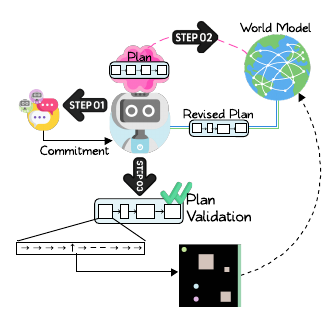}
  \caption{Workflow of DR. WELL framework. In Step 1, an agent enters the communication room with other idle agents to begin negotiation and exits with a single commitment. For instance, in a block pushing environment, the commitment may be represented by a block ID, which the agent then tries to move to the goal zone. It then reasons and generates a plan to accomplish the committed task. In Step 2, the agent refines the plan using the world model, which pushes the plan toward a more effective form. Once revised, the controller validates plans. Interaction with the environment occurs by decomposing symbolic actions into their primitive form, while the environment simultaneously updates the agent’s understanding as other agents execute their own plans.}  
  \label{fig:wholeswell}
  \vspace{-4mm}
\end{wrapfigure} 


This view follows the neurosymbolic perspective, which integrates neural methods for processing high-dimensional sensory data with symbolic reasoning for interpretability, verifiability, and compositional structure~\citep{garcez2023neurosymbolic}. Concept libraries enable agents to plan, generalize, and reuse knowledge across tasks~\citep{mao2025neuro}, separating high-level reasoning from low-level control and supporting efficient acquisition and composition of new concepts~\citep{moon2021plugin, li2022towards}. Such modularity facilitates continual learning and transfer, allowing concepts learned in one domain to be reused in another~\citep{chaudhuri2021neurosymbolic, liang2024visualpredicator}. Building on this view, we envision future agents that plan symbolically and learn from feedback grounded in environmental interaction, progressively refining their conceptual understanding of the world.

In this paper, we propose DR. WELL (\textbf{D}ynamic \textbf{R}easoning and Learning with Symbolic \textbf{W}orld Model for \textbf{E}mbodied \textbf{LL}M-Based Multi-Agent Collaboration), a decentralized neurosymbolic planning framework that enables LLM-based agents to collaborate on completing a set of interdependent tasks through a dynamic world model.
Task allocation is achieved through a structured two-phase negotiation protocol. In the proposal stage, each agent suggests a candidate task and provides a brief rationale. In the commitment stage, they converge on a joint allocation under consensus and quorum constraints. Communication is limited to these rounds, in contrast to prior “free talk” approaches~\citep{wang2023unleashing, li2023camel}, so agents synchronize only when idle.
\jes{After commitment, each agent drafts a plan from scratch with its LLM embodiment, a \textit{neural} module that interprets context and generates candidate actions. The plan is then refined using a shared symbolic world model that encodes both a description of the current environment and a graph of past events and experience.} This representation aligns agents’ reasoning without requiring disclosure of private plans. The final plan is expressed as a sequence of \textit{symbolic} actions drawn from a compact vocabulary, mediated by a controller that checks preconditions locally while the environment confirms effects.
As agents act, the outcomes are written back into the world model, enabling future planning to draw on both the current state and accumulated experience. This design preserves decentralization, limits communication to essential points of synchronization, and provides interpretability because agents justify their choices during negotiation and carry out plans composed of explicit symbolic actions.


Our main \textbf{contributions} are as follows:
\begin{itemize}
    \item We introduce a \textbf{two-phase negotiation protocol} that enables decentralized task allocation and role commitment under structured communication.
    \item We develop a \textbf{dynamic symbolic world model} that accumulates shared experience, captures reusable plan prototypes, and guides self-refinement of agent plans.
    \item We demonstrate that integrating symbolic reasoning with embodied LLM planning improves coordination efficiency and success rate in cooperative multi-agent environments.
\end{itemize}

Section~\ref{sec:methodology} presents the methodology of DR. WELL, including the negotiation protocol, symbolic planning mechanism, and the structure of the dynamic world model. Section~\ref{sec:exp_setup} describes the experimental setup and the Cooperative Push Block environment used for evaluation, along with the symbolic action vocabulary. Section~\ref{sec:exp_result} reports the experimental results, comparing DR. WELL with baseline agents and analyzing the evolution of the world model across episodes. Finally, Section~\ref{sec:conclusion} concludes the paper and discusses future directions for extending the framework.
\section{Methodology}
\label{sec:methodology}

We now detail our neuro-symbolic framework, which is designed to make coordination more tractable through symbolic abstraction and more scalable through decentralized, LLM-embodied agents. 
Agents operate in a fully observable environment where they can perceive the positions of all agents and objects and know who their teammates are, but they do not share their intended plans.

\subsection{Negotiation Protocol}
\begin{figure}
    \centering
\includegraphics[width=0.85\linewidth]{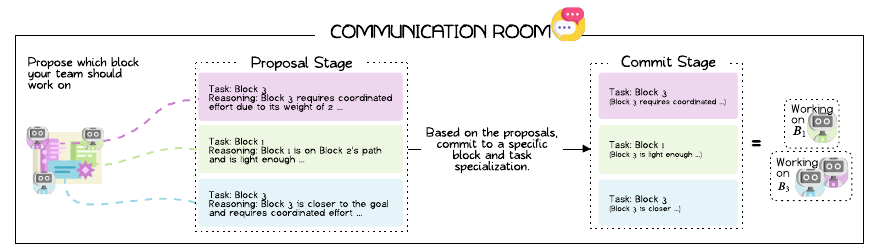}
    \caption{Two-stage Negotiation protocol: In the \textsc{proposal} stage, agents suggest candidate tasks with reasoning about feasibility and coordination needs. In the \textsc{commit} stage, they converge on a joint decision and specialize roles before independently planning over the shared world model.}
    \label{fig:negotiation}
\end{figure}

Coordination is governed by a structured two-round negotiation protocol that is triggered whenever one or more agents become idle.
The process unfolds in a round-robin order defined by agent identifiers, $\sigma_t = (a_1, \ldots, a_m)$, with all idle agents entering a shared communication room. An example of such a scenario is illustrated in Figure~\ref{fig:negotiation} for a block-pushing environment where agents try to push blocks into a given goal zone~\citep{yangcube}.

\jes{Let  $ \mathcal V_{\text{task}}$ denote the set of all symbolic task identifiers currently active in the environment (e.g., block IDs in the block-pushing domain). In the first round, each agent  $a_j\,[j=1{:}m] \in \sigma_t$ \textsc{propose}s a candidate task $\mathsf{p}_{a_j} \in \mathcal V_{\text{task}} $ along with a reasoning for its choice, based on its view of the environment.
The rationale is given in natural language, and may cite factors such as resource requirements, spatial positioning, or the potential for coordinated progress.  
This structure enforces both a discrete proposal space (task identifiers) and a free-form reasoning channel (explanations), the latter serving as the only opportunity for agents to directly communicate their justification. 
These natural language statements are intended for other LLM-based agents to observe and take into account during coordination.  
In the second round, each agent $a_j$ reviews the proposals and accompanying reasoning, then \textsc{commit}s to one task \(c_{a_j} \in \mathcal V_{\text{task}}\). Allocations are only finalized if they satisfy consensus and quorum constraints (e.g., a task requiring $k$ agents only proceeds if at least $k$ agents commit). Collectively, these individual decisions define the current agent-task mapping:
\[
M_t = \{(a_j, c_{a_j}) \mid a_j \in \mathcal{A}_t^{r}\}, \qquad M_t : \mathcal{A}_t^{r} \rightarrow \mathcal V_{\text{task}}.
\]
with $\mathcal{A}_t^{r}$ the set of agents present in the room. The mapping \(M_t\) records the commitments and concludes the negotiation phase.}

This two-round design ensures that proposals are explored before commitments are made, reducing conflicts and deadlocks in decentralized coordination.  
The negotiation protocol is asynchronous. 
If only a single agent is available, it may wait or select a solo-feasible task. 
When multiple agents are free, they resynchronize and negotiate again. 
This produces a natural cycle of synchronization, execution, asynchrony, and resynchronization, a pattern characteristic of embodied cooperation. 
Once commitments are finalized, each agent knows its teammates for the next task, but the content of individual plans is not shared. \jes{A complete symbolic specification of the negotiation protocol appears in Appendix~\ref{formalization}.}

\subsection{Symbolic Planning and Execution}

Following task commitment, at every (re-)planning point, each agent, independently, generates a plan from scratch through its LLM embodiment.  
The agent’s commitment is encoded into the LLM prompt, ensuring the plan is grounded in the agreed-upon task.
The draft is then revised using the shared world model (WM), which combines a symbolic projection of the current environment with a graph of past events and experience (details in Section~\ref{worldmodel}).  
As the best source of guidance available to the agents, the WM pushes the draft toward a more effective and context-aware form.  
The revised plan takes the form of a sequence of symbolic macro-actions drawn from a fixed vocabulary (specified in Section~\ref{symbolic}), with their order and content determined by its reasoning under the environment rules and goals,  to the best of its knowledge.  
This output is then validated to ensure only admissible actions appear. 
Execution proceeds under a \textsc{plan → execute → re-plan} cycle, where the symbolic controller mediates step-by-step execution and triggers re-planning as needed. This process is depicted in Figure~\ref{fig:wholeswell} from the perspective of a single agent, in order to highlight the decentralized characteristics of the proposed methodology.

The controller \textsc{(i)} checks whether the pre-conditions of each symbolic action are satisfied in the current symbolic world model, \textsc{(ii)} translates symbolic steps into primitive moves, and \textsc{(iii)} advances execution as a state machine. 
Post-conditions, however, are verified by the environment: once primitive moves are applied, the environment reports whether the intended effects were realized. 
If a precondition is not satisfied, e.g., a cooperative action requires more agents than are aligned, the controller places the agent in a waiting state with a timeout counter. 
If quorum is not reached before the timeout expires, the action is marked as failed and the plan index is advanced. 
If no further steps remain, the agent transitions back to {idle} and re-plans. 
When one or more agents become idle and enter the communication room, the environment pauses, suspending agents still in the middle of executing their plans until the idle agents reach a new commitment and exit the communication room.
This division ensures that the controller enforces symbolic logic locally, while the environment provides ground-truth confirmation of effects, yielding continuous progress without deadlock.

\begin{figure}
    \centering
    \includegraphics[width=0.88\linewidth]{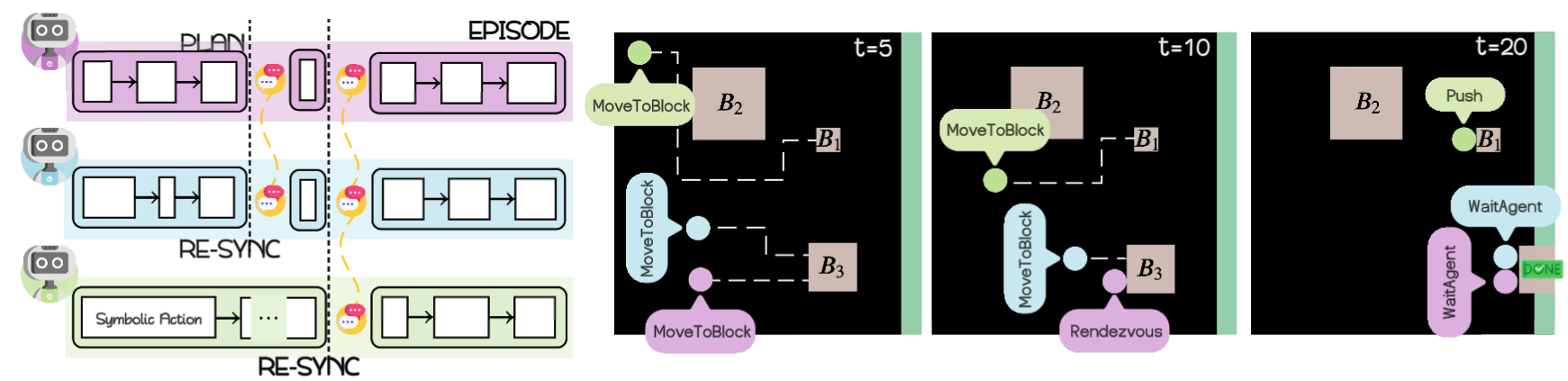}
    \caption{Post-commitment planning and execution cycle in a block-pushing environment. \jes{Each color represents a different agent, showing its planning and execution timeline.  The speech bubble  marks the communication room where agents synchronize and negotiate before proceeding with execution.
    After consensus, each agent expands its assigned role into a full symbolic plan and executes it independently. These plans are shown on the left as sequences of white rectangles, each representing a symbolic action (e.g., \textsc{Push}), whose varying widths indicate different durations.
    Once a plan is completed, the agent re-enters the communication room, the environment then pauses until all idle agents resynchronize, briefly suspending those still mid-plan. 
 On the right, snapshots at different timesteps ($t=5,10,20$) show how agents coordinate to approach and push blocks toward the goal zone, where completed blocks are marked \texttt{DONE} and excluded from future negotiations. }}
    \label{fig:planedenv}
    \vspace{-5mm}
\end{figure}

\subsection{World Model} \label{worldmodel}

To support planning and coordination without explicit plan sharing, agents rely on a shared symbolic world model (WM). 
The WM serves as the agents’ model of the world, providing a continuously updated symbolic representation of environment state, actions, and outcomes that grounds coordination and reasoning.

It encodes the environment state in symbolic form for \textit{planning} and \textit{logging}, while the environment itself remains the authoritative ground truth.
At every timestep, the WM records a snapshot of the world, including the positions and statuses of all agents, the state of all objects, the symbolic action being executed with its temporal bounds, and each agent’s progress through its plan. 
Agents only observe the structured environment state, positions, availability signals, and object statuses, without privileged access to the detailed plan steps of others. The full trace is retained in the WM and can be post-processed into timeline visualizations that highlight idle periods, bottlenecks, and patterns of coordination.

Having outlined the general structure of the WM, we next specify the information it provides in two phases of the DR. WELL framework: \textsc{(i)} negotiation and \textsc{(ii)} planning.

\paragraph{\textsc{(i)} Negotiation guidebook.} During negotiation, the world model provides agents with structured context about the current session and prior experience. This includes the current timestep, the number of active agents, summaries of how tasks have historically performed (e.g., frequency of attempts, completion rates, and average durations), and estimates of effective team sizes. These statistics represent aggregated knowledge about past coordination attempts, giving agents a common reference point when proposing and evaluating tasks. By grounding proposals in shared historical evidence, the world model helps agents reason about which tasks are promising, which may be infeasible, and how to align commitments with available teammates. The full trace is provided in Appendix~\ref{WM}.

\begin{center}
\begin{tcolorbox}[
  title={\includegraphics[height=1.5em]{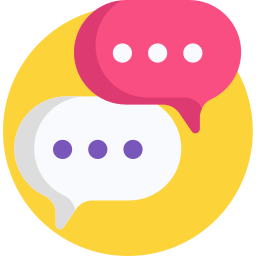} \,Phase 1: Communication — Example Output},
  width=0.99\linewidth,       
  breakable,
  colback=myyellow!6,            
  colframe=black,            
  colbacktitle=myyellow!20,      
  coltitle=black,            
  fonttitle=\bfseries,
  arc=2mm,                   
  boxrule=0.6pt,
  left=6pt,right=6pt,top=6pt,bottom=6pt,
]
\footnotesize

\textbf{CURRENT SESSION INFO}:
\hspace*{1em} \texttt{Current timestep:[T]},
\hspace*{1em} \texttt{Number of agents:[N]}

 \vspace*{3mm}\textbf{HISTORICAL TASK PERFORMANCE}
\\
\hspace*{1em}\texttt{Task\_A: avg\_start=[t], range=[start-end], success\_rate=[\%] (/[\# attempts])} \\
\hspace*{1em} \texttt{Task\_B: ...}

 \vspace*{3mm}\textbf{OPTIMAL TEAM SIZE RECOMMENDATIONS }

  \hspace*{1em} \texttt{Task\_A: optimal team size = [\#/UNKNOWN] (success\_rate$_{\texttt{best}}$:[\%])} \\ 
  \hspace*{1em} \texttt{Task\_B: ...}

\end{tcolorbox}
\end{center}


\paragraph{\textsc{(ii)} Plan library.}
After commitment, the world model supports plan construction in two steps. First, it provides \textit{plan prototypes}, abstract sequences of symbolic operations, that have been attempted for \textit{similar} tasks in the past, sorted by their observed success rates and efficiency. These prototypes give each agent a starting point for generating a candidate plan. 
Second, the model supplies detailed \textit{plan instances} that capture concrete instantiations of these prototypes, along with metadata such as success rate, number of attempts, and execution time. 
Agents use this information to refine their own plan drafts, balancing prior knowledge with their current commitments.
Together, this organization enables agents to assess whether a plan used in a comparable context is likely to succeed and to refine their own drafts by drawing on alternatives with better performance. 
A complete version of the trace can be found in Appendix~\ref{WM}.

\jes{Together, these uses show how the world model implicitly links negotiation and planning into a continuous process of coordination.
}
In doing so, the world model functions as both a shared memory and a predictive scaffold that guides agents toward increasingly effective collaboration.
\jes{While the symbolic actions themselves are domain-specific, the underlying mechanism for storing and evaluating structured plans is domain-agnostic and applicable to any setting that supports symbolic representations of action and state.
}

\begin{tcolorbox}[
  title={\includegraphics[height=1.6em]{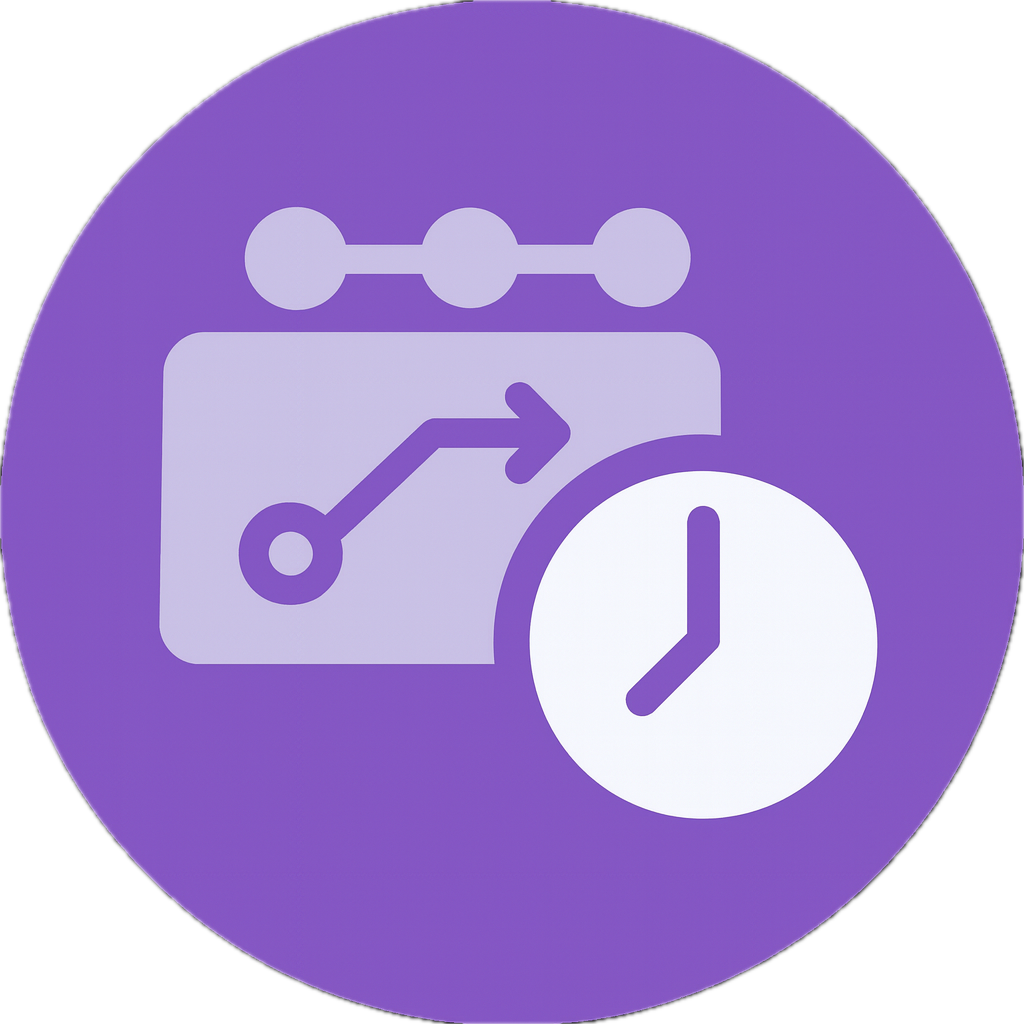} \,Phase 2: Planning — Example Output},
  width=0.99\linewidth,       
  breakable,
  colback=mpurple!6,            
  colframe=black,            
  colbacktitle=mpurple!25,      
  coltitle=black,            
  fonttitle=\bfseries,
  arc=2mm,                   
  boxrule=0.6pt,
  left=6pt,right=6pt,top=6pt,bottom=6pt,
]
\footnotesize
\centerline{--- PROTOTYPES ---}
\vspace*{1mm}
{Planning for Task\_X (example)...}
\vspace*{2mm}

\textbf{Historical Plan Prototypes (ranked by success rate):}
 {\small
 \vspace*{2mm}
 
  \hspace*{1em}\texttt{1. Success rate=[\%] | avg team=[\#] | avg duration=[steps/UNKNOWN]} \\ 
    \hspace*{3.5em} \texttt{Prototype:[SymAct $\rightarrow$ SymbAct$'$ $\rightarrow$ SymAct$''$]}

     \vspace*{1mm}
  \hspace*{1em} \texttt{2. ...}}

\vspace*{3mm}

\centerline{--- INSTANCES ---}
\vspace*{2mm}

\textbf{Detailed Plan Instances (ranked by success then duration):}
\vspace*{2mm}
  {\small
  
  \hspace*{1em}\texttt{1. Success rate=[\%] | attempts=[\#] | duration=[steps/UNKNOWN]}
  \\
     \hspace*{3.5em} \texttt{Plan:[SymAct(param) $\rightarrow$ SymbAct$'$(param) $\rightarrow$ SymAct$''$(param)]}

      \vspace*{1mm}
  \hspace*{1em} \texttt{2. ...}}

\end{tcolorbox}

\begin{figure}
    \centering
    \includegraphics[width=.75\linewidth]{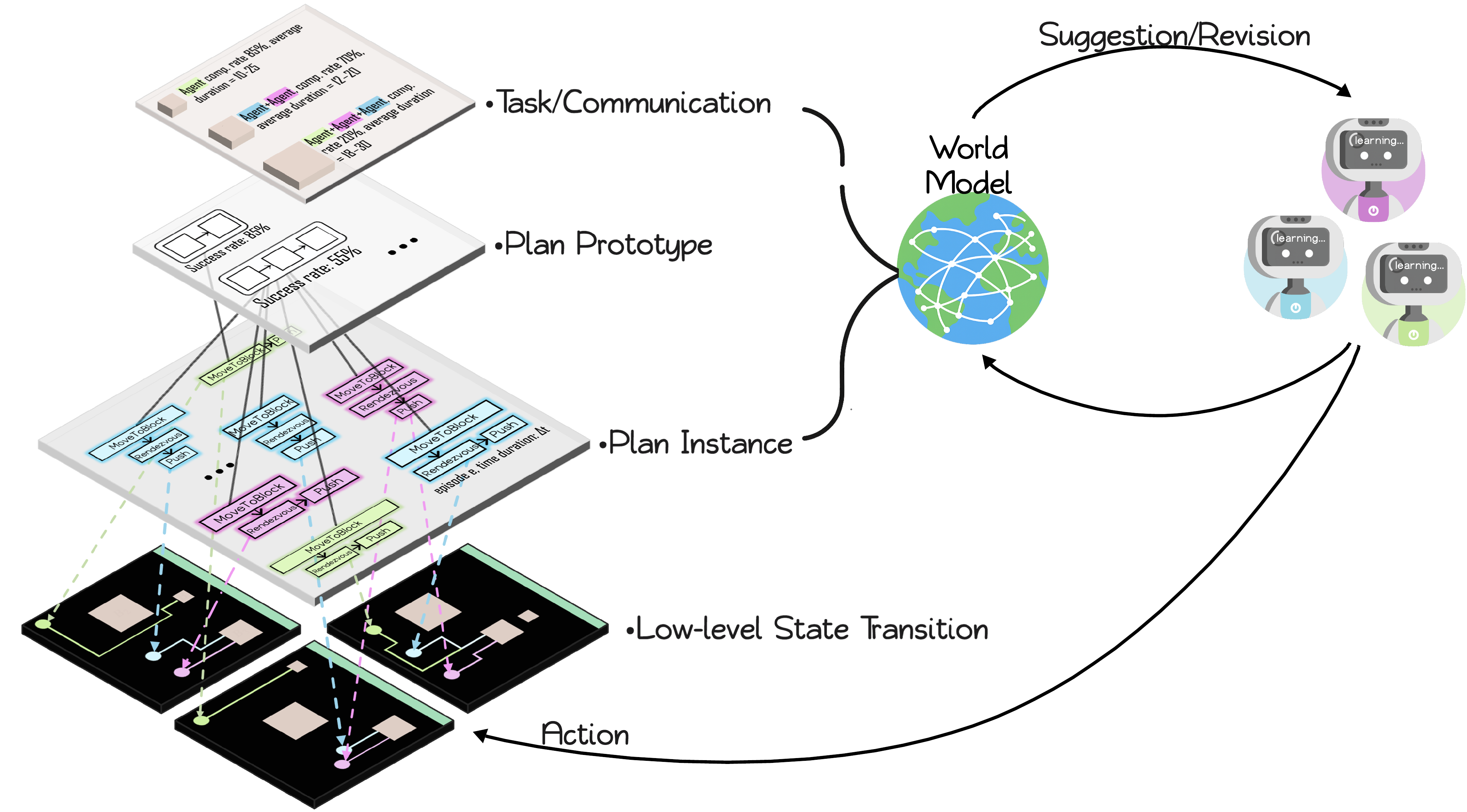}
    \caption{View of the world model (WM), showing how it organizes knowledge across multiple layers: task and communication history, plan prototypes, and detailed plan instances. These layers acts as a shared memory that aggregates state and experience, ensuring agents draw on the same knowledge base while keeping execution decentralized and private. }
    \label{fig:layersofWM}
\end{figure}

\subsubsection{Multi-Episode Graph Construction from World Traces}

Earlier (in Figure~\ref{fig:wholeswell}), we illustrated how individual agents interact with the world model, showing how it is queried and updated within each agent’s planning cycle. 
Figure~\ref{fig:layersofWM} expands this view, revealing that the world model is itself structured as a dynamic symbolic graph that organizes knowledge across multiple layers. Formally, the world model is represented as
$$\mathcal{G = (V,E)}, \quad
\mathcal{V = V_{\text{epi}} \cup V_{\text{task}} \cup V_{\text{proto}} \cup V_{\text{inst}}, \quad
E = E_{\text{epi}\rightarrow\text{task}} \cup E_{\text{task}\rightarrow\text{proto}} \cup E_{\text{proto}\rightarrow\text{inst}}},$$
where $\mathcal{V_{\text{epi}}, V_{\text{task}}, V_{\text{proto}}, \text{ and } V_{\text{inst}}}$ correspond to episodes, tasks (block identifiers), plan prototypes (argument-free symbolic sequences), and plan instances (prototypes with instantiated parameters), respectively. Edges $\mathcal{E}$ capture the hierarchical relations among these layers. Each new episode contributes a temporal trace of task allocations, plans, and symbolic actions (Figure~\ref{fig:world-trace}), which is integrated into this evolving graph. The resulting representation (Figure~\ref{fig:dynamic-world}) positions higher-level abstractions, episodes and tasks, near the center, with prototypes and instances extending outward. Each plan instance $v \in \mathcal{V_{\text{inst}}}$ is grounded to its observed outcome $o(v)\!\in\!\{0,1\}$, allowing upper-level nodes to aggregate statistics from their descendants and thereby capture evolving patterns of coordination and strategy effectiveness across episodes.

\textbf{Graph Update.}
The variable $k$ indexes points at which the world model is updated; in this work, the WM is updated episodically. Each episode produces an additive subgraph $\Delta\mathcal{G}_k=(\mathcal V_k^{+}, \mathcal E_k^{+})$, and the model evolves as
${\mathcal G_{k+1} = \mathcal G_k \cup  \Delta \mathcal G_k.}$

\section{Experimental Setup} \label{sec:exp_setup}
\subsection{Environment: Cooperative Push Block}
We evaluate our framework in a customized \textsc{Cooperative Push Block} environment, termed CUBE~\citep{yangcube}, built on PettingZoo’s parallel API~\citep{terry2021pettingzoo}. CUBE mirrors the structure of the Unity ML-Agents Cooperative Push Block task~\citep{juliani2020}, while extending it with the components necessary to support the learning of embodied LLM-based agents.
The environment is a grid world in which agents must coordinate to move blocks of varying sizes into a shared goal zone located along one side of the grid. Each block is represented as a $w \times w$ square with weight $w$. \jes{Moving a block of weight $w$ requires $w$ agents to push simultaneously on the same face, causing the block to move one cell in the direction of the push.} Thus, the level of cooperation is directly tied to block size: small blocks can be moved individually, while larger blocks ($w \geq 2$) require collaboration among multiple agents. The objective is to push blocks into the goal zone using the fewest possible steps.

Observations are provided in two modalities: a multi-channel tensor and a symbolic representation. The tensor encodes spatial features, with separate channels for agent positions, block locations, and block weights. The symbolic representation specifies each agent’s identifier and position, as well as each block’s size, position, and distance to the goal.  
Agents act by selecting from a symbolic action space, where each symbolic action is instantiated as a sequence of primitive actions executed in the embodied environment under physical and spatial constraints. The task requires agents to interpret their observations, recognize the level of cooperation needed for each block, and coordinate their actions to maximize the total group reward.

\subsection{Symbolic Actions} ~\label{symbolic}
The symbolic action vocabulary is tailored to the dynamics of the \textsc{Cooperative Push Block} environment. Each action abstracts the minimal set of operations agents require to approach, align, synchronize, and collectively move heavy blocks. By capturing these primitives, we provide just enough structure to enable coordination while keeping the action space compact and interpretable.


\begin{itemize}[leftmargin=*, itemsep=0.1ex, topsep=0.0ex]
    \item \textsc{WaitAgents(count = $k$, timeout =  $t$)}: wait until $k$ agents are idle, or until timeout $t$ expires
    \item \textsc{Rendezvous(block$_{id}$, side, count = $k$, timeout = $t$)}: wait until $k$ agents arrive to the specified \textsc{side} of the block, or until timeout $t$ expires
    \item \textsc{MoveToBlock(block$_{id}$, side)}: move and align to the specified \textsc{side} of the block
    \item \textsc{Push(block$_{id}$, steps = $n$)}: push the target block for $n$ steps if aligned, otherwise remain idle
    \item \textsc{YieldFace(block$_{id}$, steps = $n$)}: move $n$ steps away from the block, opposite its facing side (inferred as opposite to the block’s approach direction)
\end{itemize}

\textbf{Strategic Consideration.}  
Although the symbolic vocabulary appears compact, the inclusion of arguments (e.g., block identifiers, sides, timeouts, and step counts) allows each action to be instantiated in many different ways.  
When expanded, these parameterized actions create a much larger planning space, making the effective action set much more expressive than it first appears.  
Within this space, agents must decide how to align with a block, synchronize with teammates, wait for collaborators, or yield when obstructing others, even when already committed to a plan.  
For instance, \textsc{MoveToBlock} is often followed by \textsc{Rendezvous} to ensure multiple agents are present at the same block face before executing \textsc{Push}.  
Such combinations highlight how flexible use of symbolic actions gives rise to complex interaction patterns and challenging coordination dynamics in multi-agent settings.





\section{Experimental Results}\label{sec:exp_result}
This section details how the world model is formed and  built up across episodes. We also use this representation to compare the baseline and our proposed agents to evaluate performance.

\subsection{World Trace}
When agents interact with the environment, their activities are recorded in a \emph{world trace}. This trace logs the tasks agents committed, the plans they proposed, and the symbolic actions they executed. It also includes the communication exchanged among agents during the process.

\begin{figure}[H]
    \centering
    \includegraphics[width=1\linewidth]{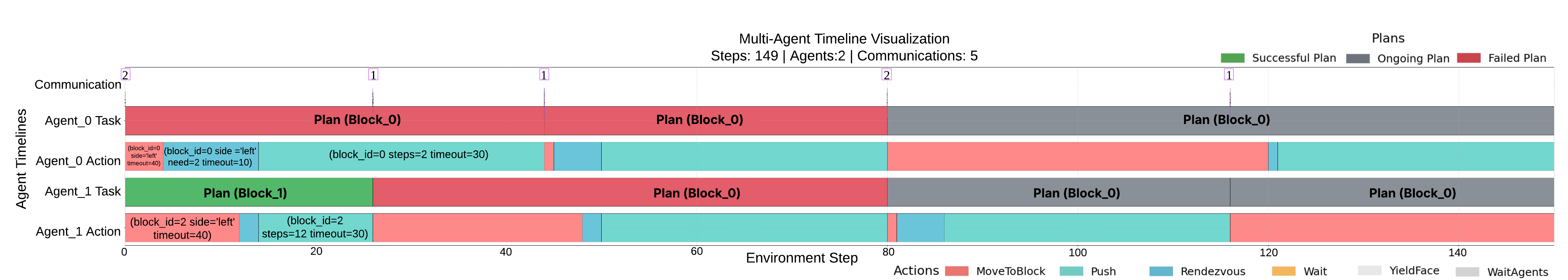}
    \caption{Two-agent game with a limit of 150 \texttt{max\_steps}. Communication events are marked along the top of the timeline. For each agent, timelines record task allocations, outcomes, and the sequence of symbolic actions (e.g., \textsc{MoveToBlock}, \textsc{Push}, \textsc{Rendezvous}).}
    \label{fig:world-trace}
\end{figure}

\textbf{Communication Timeline.}
Communication drives task allocation and coordination. As shown in the top row of Figure~\ref{fig:world-trace}, the set of participating agents varies over time, reflecting changing task commitments and points where negotiation or synchronization is required. In this episode, five exchanges occur, each producing a spike aligned with the onset of a new plan.

\textbf{Task Timeline.}
Tasks are symbolically represented using discrete environment structures (blocks) to simplify communication. We also define a symbolic concept for obtaining feedback on task success, implemented as checking whether the block remains in the environment. This provides a uniform way to evaluate whether a task has been achieved.
In Figure~\ref{fig:world-trace}, these commitments appear in the \emph{Plan Timeline} lanes of each agent. For example, Agent~0 initially commits to a plan that ultimately fails (red), while Agent~1 pursues a longer-lasting successful plan (green). At the end of each plan, the model checks whether the target block still exists; if not, the task is marked as completed. This evaluation links agents’ symbolic commitments with concrete outcomes, closing the loop between planning and execution.

\textbf{Symbolic Actions Timeline.}
The \emph{Action} lanes in Figure~\ref{fig:world-trace} show how high-level plans unfold into symbolic actions. A given task can be realized through different action patterns, and even when the same pattern is executed, outcomes may differ depending on the evolving state and uncertainty of the embodied environment.

\paragraph{Multi-agent Timeline.}
The world trace records each agent’s behavior over an episode, making visible how intentions, actions, and outcomes align to support multi-agent collaboration.
Figure~\ref{fig:world-trace} illustrates both \textit{synchronization} and \textit{divergence} between the two agents. At some points agents divide work across tasks, at others they collaborate on the same task, and before engaging with a block they explicitly synchronize through \textsc{Rendezvous} actions. 
Agents periodically reconnect through communication, during which new task allocation strategies are negotiated and commitments are updated. In this way, the multi-agent relationship is expressed both through \emph{intention alignment}, reflected in the distribution of tasks across agents, and through \emph{spatio-temporal grounding}, captured in the alignment of their plans and actions. This interplay between independent execution and coordinated synchronization is central to effective multi-agent collaboration. 

\subsection{Dynamic Symbolic World Model}
\begin{figure}[htbp]
    \centering
    \begin{subfigure}[b]{0.32\linewidth}
        \centering
        \includegraphics[width=\linewidth]{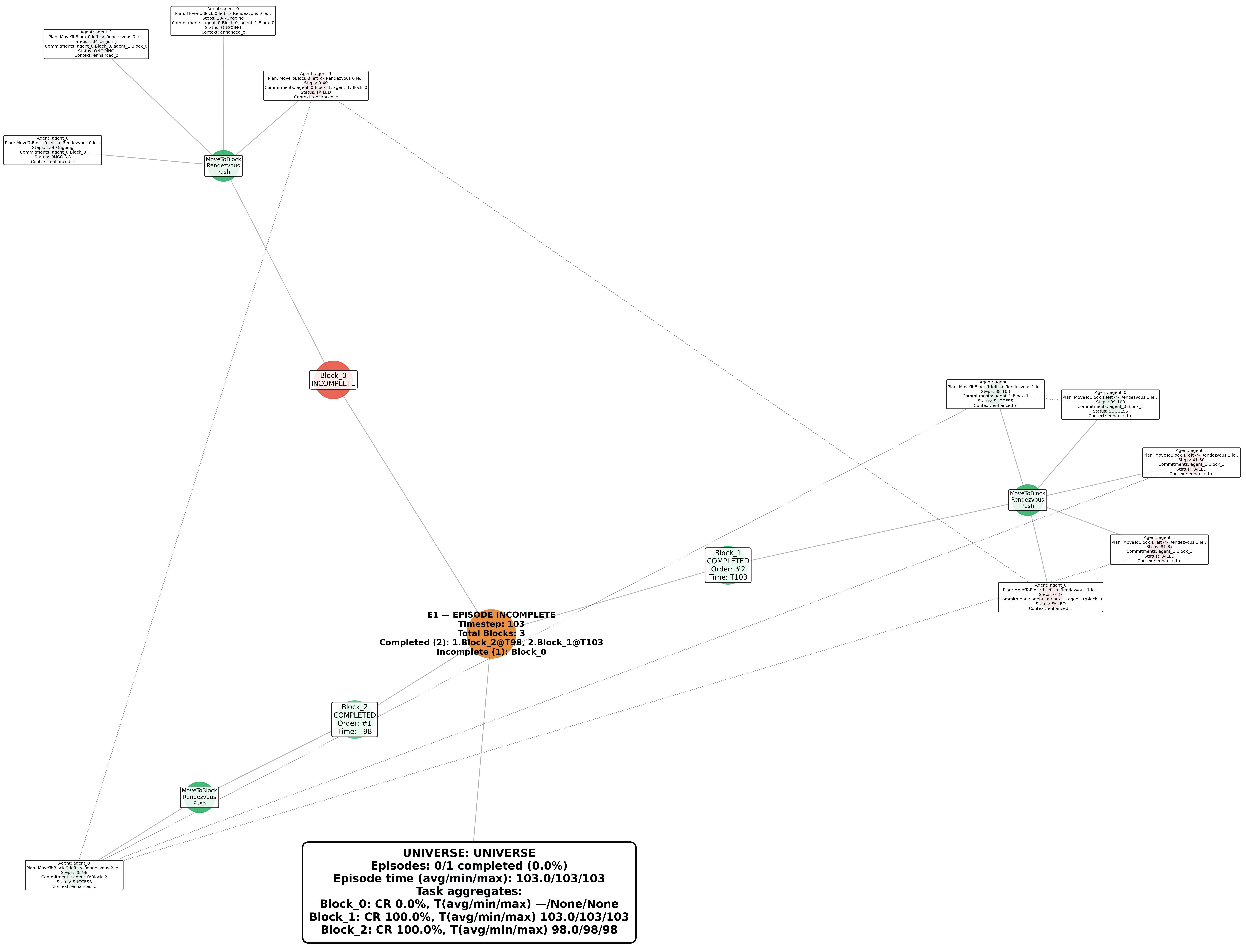}
        \caption{\hyperref[fig:E1]{Episode 1}}
        \label{fig:world_model_graph_ep1}
    \end{subfigure}
    \hfill
    \begin{subfigure}[b]{0.32\linewidth}
        \centering
        \includegraphics[width=\linewidth]{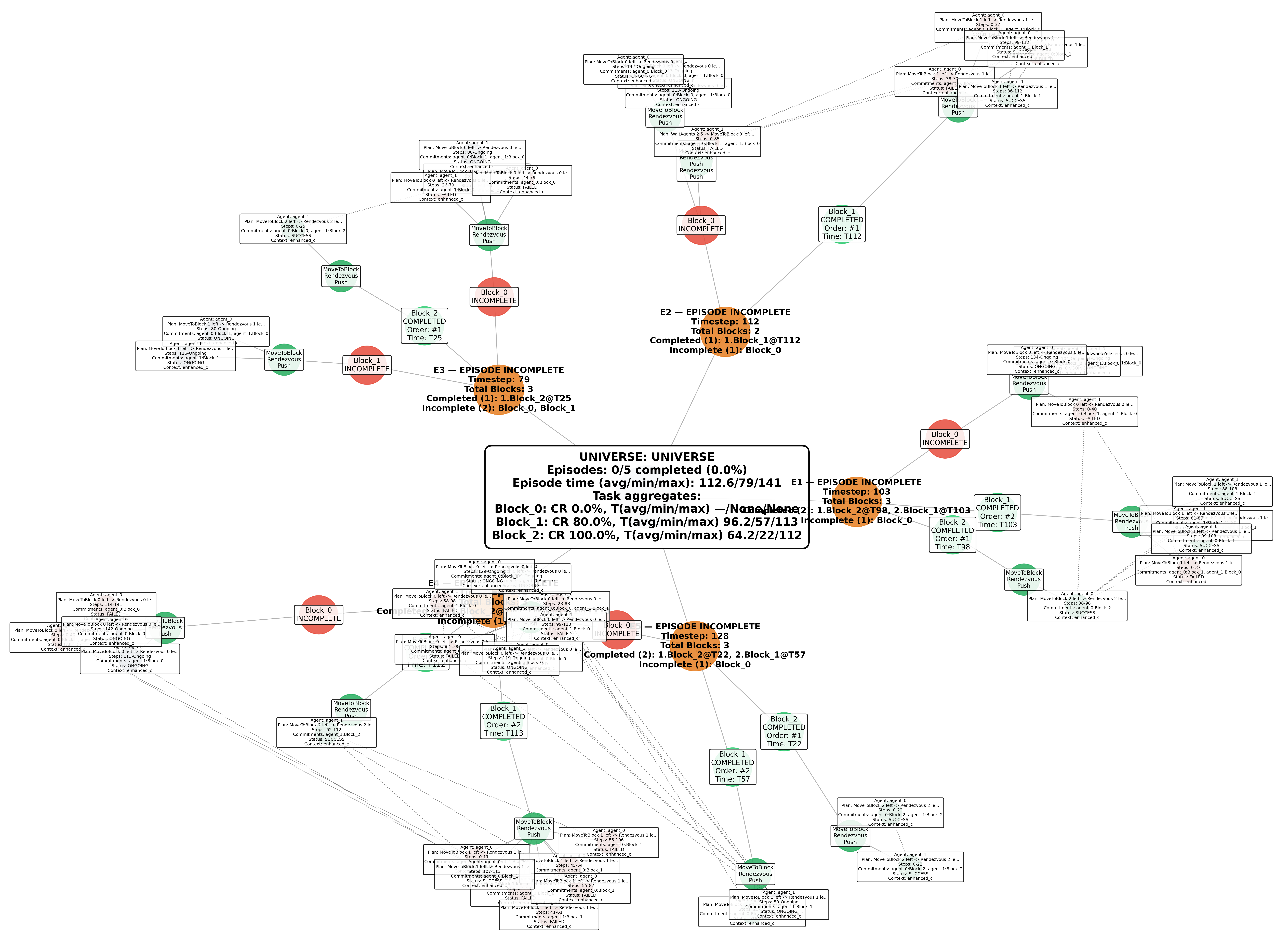}
        \caption{\hyperref[fig:E5]{Episode 5}}
        \label{fig:world_model_graph_ep5}
    \end{subfigure}
    \hfill
    \begin{subfigure}[b]{0.32\linewidth}
        \centering
        \includegraphics[width=\linewidth]{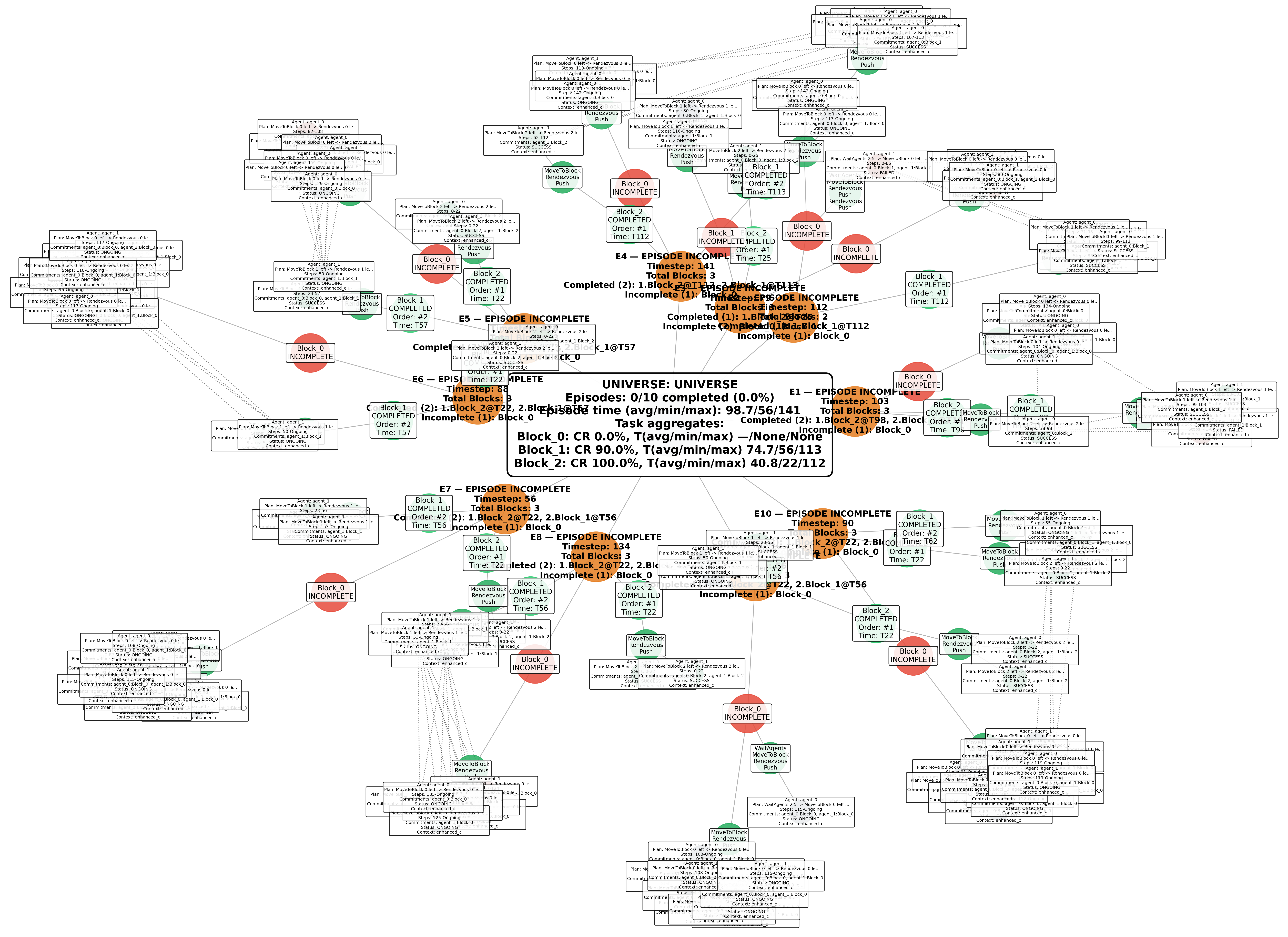}
        \caption{\hyperref[fig:E10]{Episode 10}}
        \label{fig:world_model_graph_ep10}
    \end{subfigure}
    
    \caption{Evolution of the Dynamic Symbolic World Model (WM) across episodes. 
Panels (a), (b), and (c) show snapshots after Episode~1, Episode~5, and Episode~10, respectively. 
Each graph instantiates the layered WM in Figure~\ref{fig:layersofWM}: episode status (center) $\rightarrow$ tasks/communications $\rightarrow$ plan prototypes $\rightarrow$ plan instances. 
Nodes are color-coded by outcome: \textcolor{mgreen}{successful task/instance [green]}, \textcolor{mred}{block incomplete [red]}, \textcolor{morange}{episode incomplete [orange]} (i.e., at least one block undelivered). 
Edges indicate the relationships from the episode summary to task nodes, from tasks to their plan prototypes, and from prototypes to concrete plan instances. 
Text inside the rectangular nodes (not legible at this scale) records plan steps, parameters, attempts, success rates, and durations. 
The intent of this figure is to show how the WM grows and becomes structurally richer over time, not to read individual labels. 
Full-resolution versions with readable text are provided in Appendix~\ref{WM_graph}.}
    \label{fig:dynamic-world}
\end{figure}

We investigate the dynamic nature of the symbolic world model by tracking how its graph expands across episodes, as illustrated in Figure~\ref{fig:dynamic-world}.
At Episode 1, the graph is sparse, containing only a few task and plan instance nodes. 
By Episode 5, clearer structure emerges, with task concepts linked to recurring plan patterns. 
By Episode 10, the graph has become dense, connecting task concepts, reusable plan templates, and detailed execution records. 

These results indicate that the world model does more than store isolated traces. It develops a symbolic memory that unifies abstractions with concrete outcomes. As episodes progress, the model captures reusable patterns and aggregates statistics such as success rates, which in turn support more effective planning and coordination. This evolving structure can be integrated into LLMs, enhancing their ability to reason, plan, and collaborate.

\subsection{Baseline Agents}
\begin{figure}[h]
  \centering
  \begin{subfigure}[b]{0.31\linewidth}
    \includegraphics[width=\linewidth]{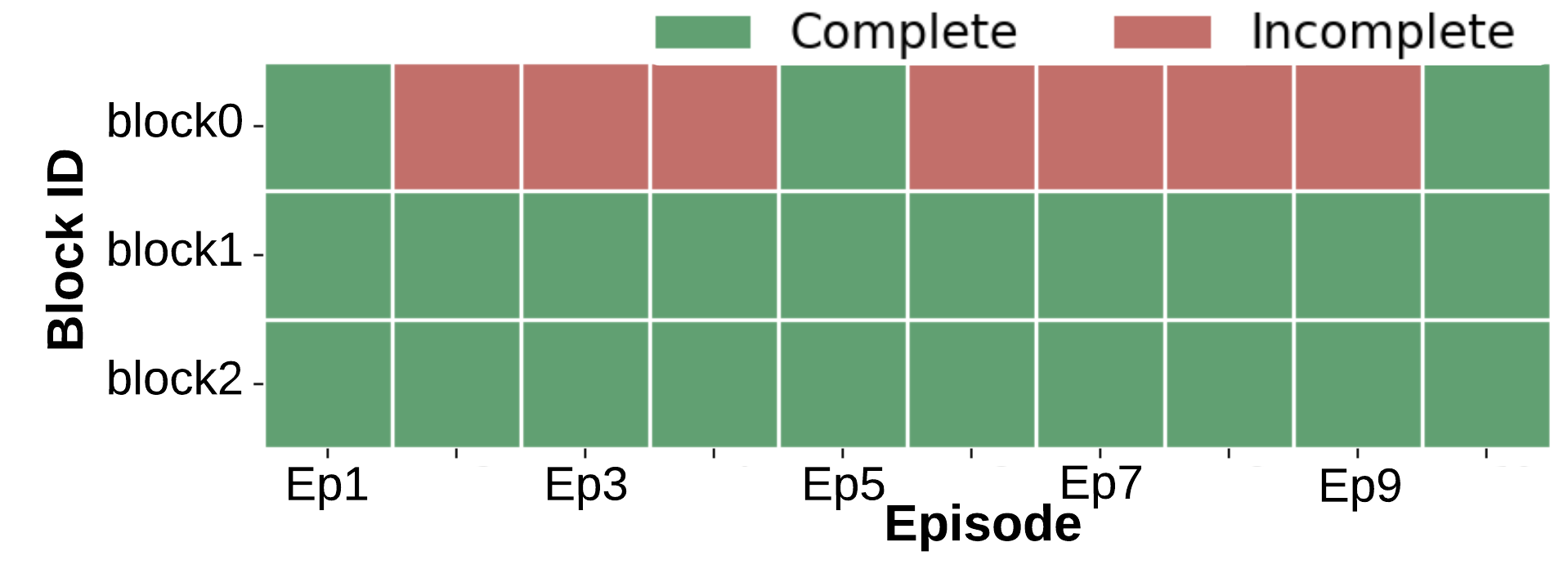}
    \caption{Block completion outcome.}
    \label{fig:block_completion}
  \end{subfigure}\hfill
  \begin{subfigure}[b]{0.32\linewidth}
    \includegraphics[width=\linewidth]{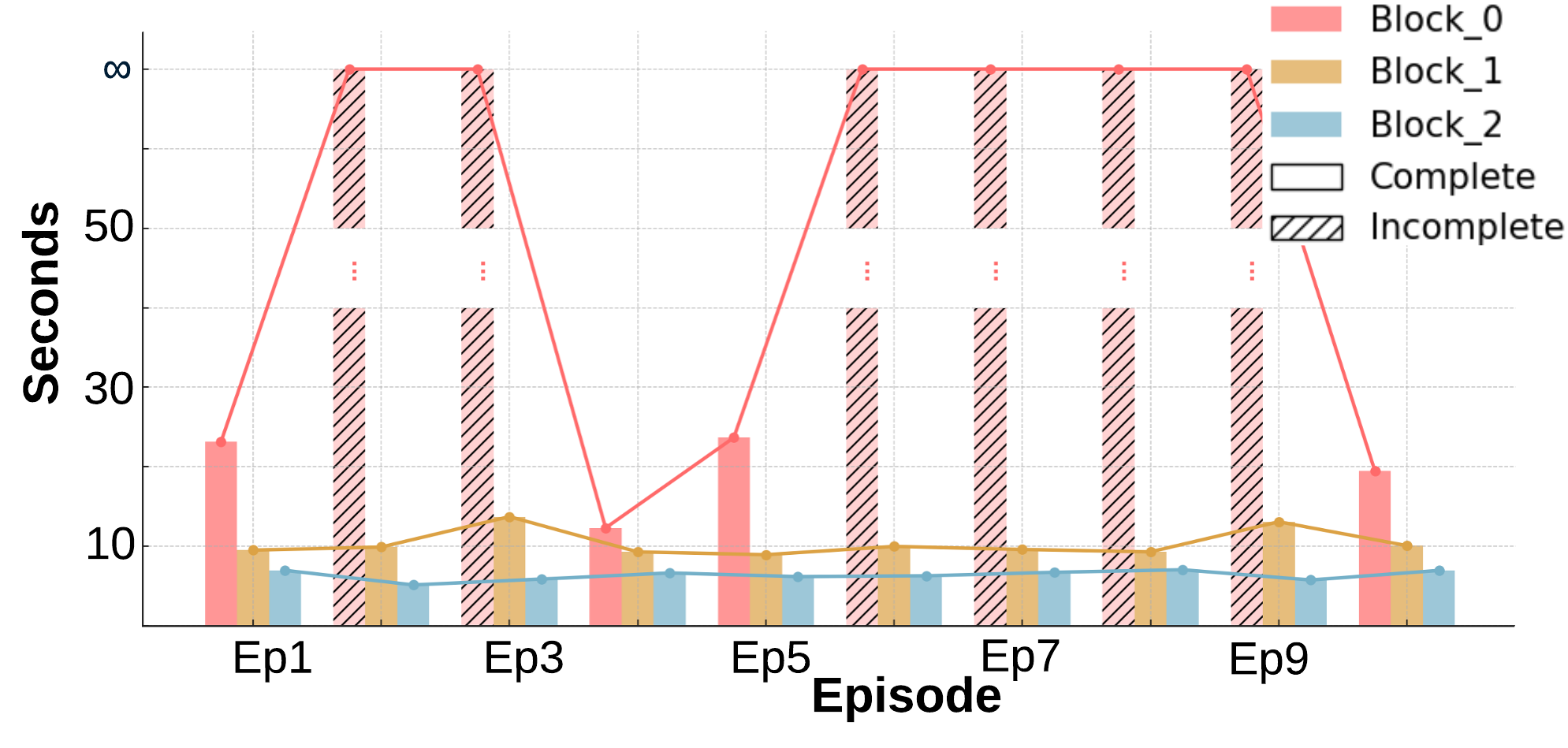}
    \caption{Completion time (Seconds).}
    \label{fig:block_timing}
  \end{subfigure}\hfill
  \begin{subfigure}[b]{0.32\linewidth}
    \includegraphics[width=\linewidth]{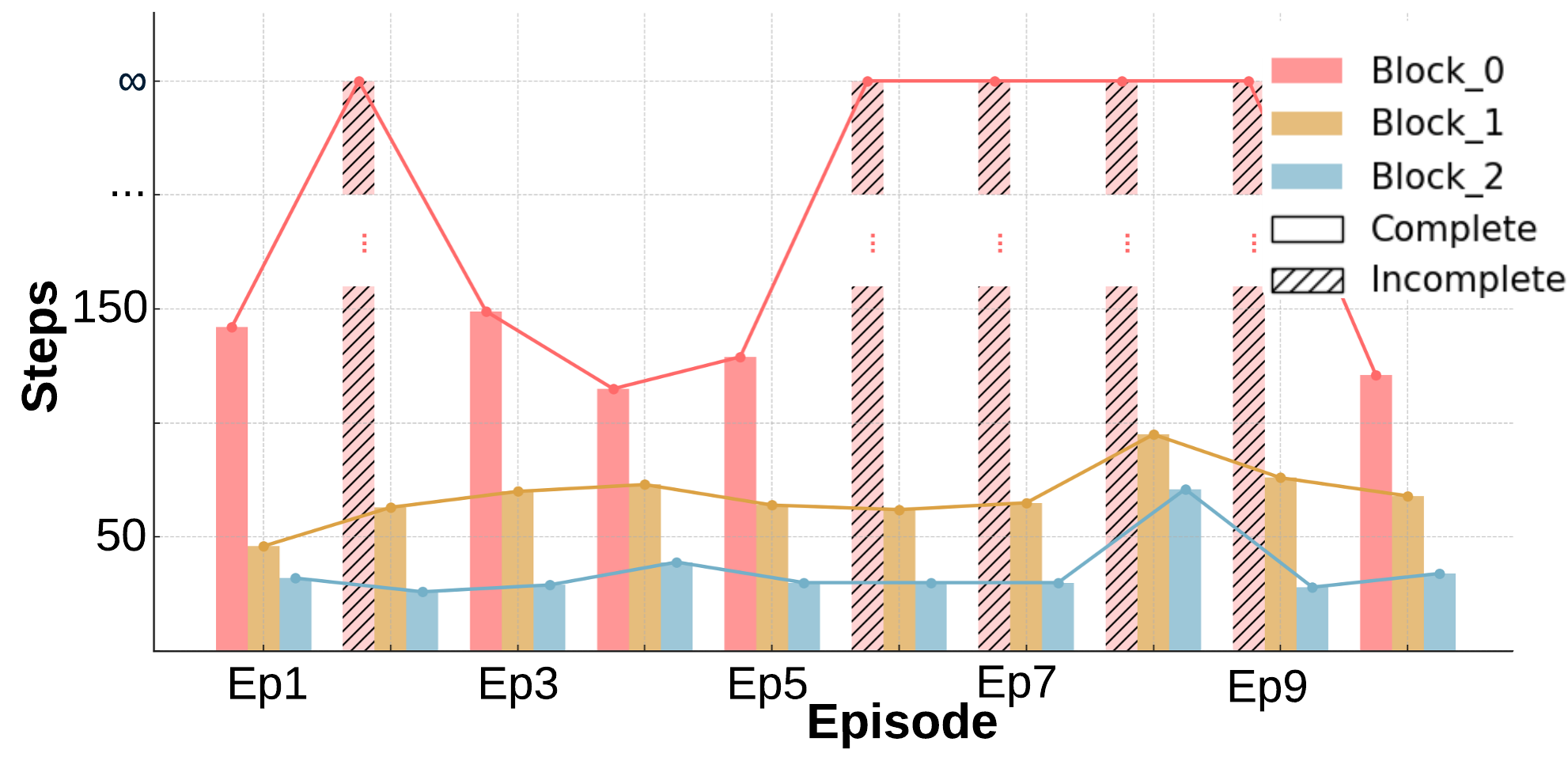}
    \caption{Completion time (Env-steps).}
    \label{fig:block_envsteps}
  \end{subfigure}
  \caption{Performance of baseline agent:
(a) Block-level completion outcomes show a binary pattern of success or failure for each block. 
(b) Completion times in wall-clock seconds and (c) environment steps demonstrate that durations remain nearly constant across episodes. 
The plots together indicate the baseline’s consistent but inflexible behavior, completing some blocks reliably while repeatedly leaving others unfinished.}
  \label{fig:baseline-results}
\end{figure}

The baseline agent operates in a purely zero-shot fashion, relying only on symbolic representation of the state and a fixed planning prompt. Agents do not negotiate tasks, make commitments, revise plans, or communicate with one another. Instead, each agent repeatedly generates short plans directly from the prompt, which encodes a simple strategy: always work with the block closest to the goal zone. This design emphasizes independent decision-making without shared memory or coordination mechanisms.

\textbf{Baseline Results.}
Figure~\ref{fig:baseline-results} shows the outcomes of the baseline agent. The block completion outcome (Figure~\ref{fig:baseline-results}a) shows a binary pattern of completed or missed blocks across episodes, without signs of learning or adaptation. The completion time plots (Figures~\ref{fig:baseline-results}b and \ref{fig:baseline-results}c) also remain nearly constant, indicating that the agents rely on a fixed policy rather than adapting over time. While agents often succeed in pushing blocks already near the goal zone, heavier or less accessible blocks are frequently left unfinished. Moreover, their task allocation strategy is inefficient: all agents tend to work on the same block, even when it does not require multiple agents’ efforts, leading to wasted efficiency.

\subsection{Our Framework Results: DR. WELL}
\begin{figure}[h]
  \centering
  \begin{subfigure}[b]{0.31\linewidth}
    \includegraphics[width=\linewidth]{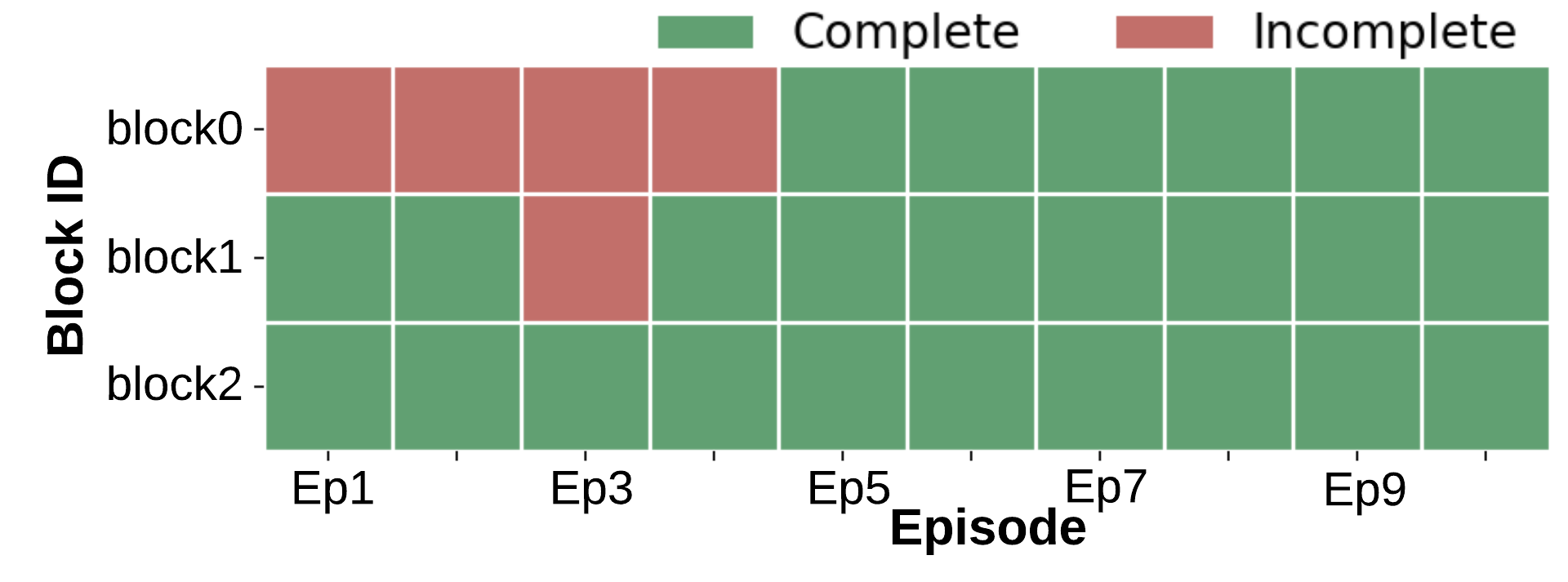}
    \caption{Block completion outcome.}
    \label{fig:block_completion}
  \end{subfigure}\hfill
  \begin{subfigure}[b]{0.32\linewidth}
    \includegraphics[width=\linewidth]{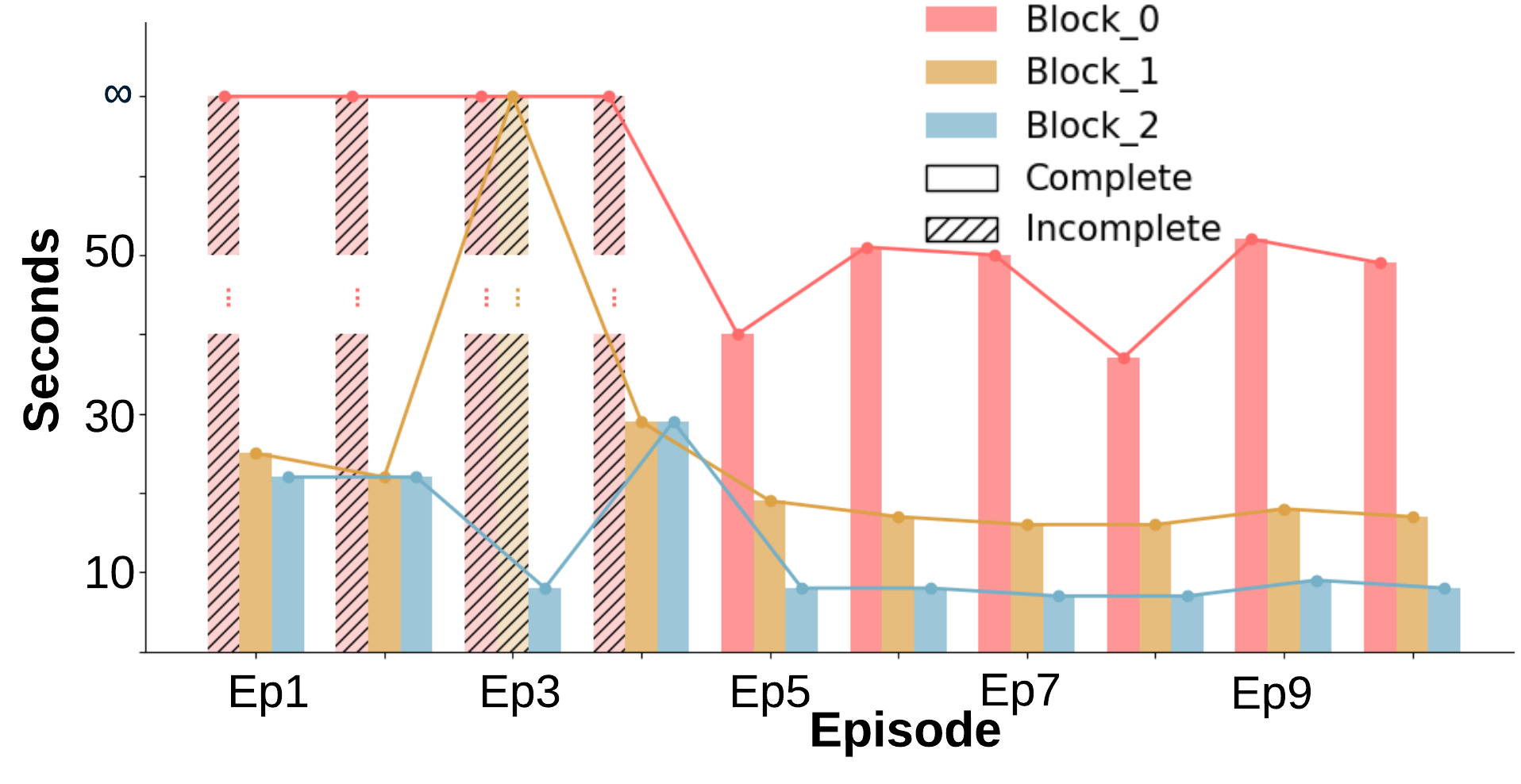}
    \caption{Completion time (Seconds).}
    \label{fig:block_timing}
  \end{subfigure}\hfill
  \begin{subfigure}[b]{0.32\linewidth}
    \includegraphics[width=\linewidth]{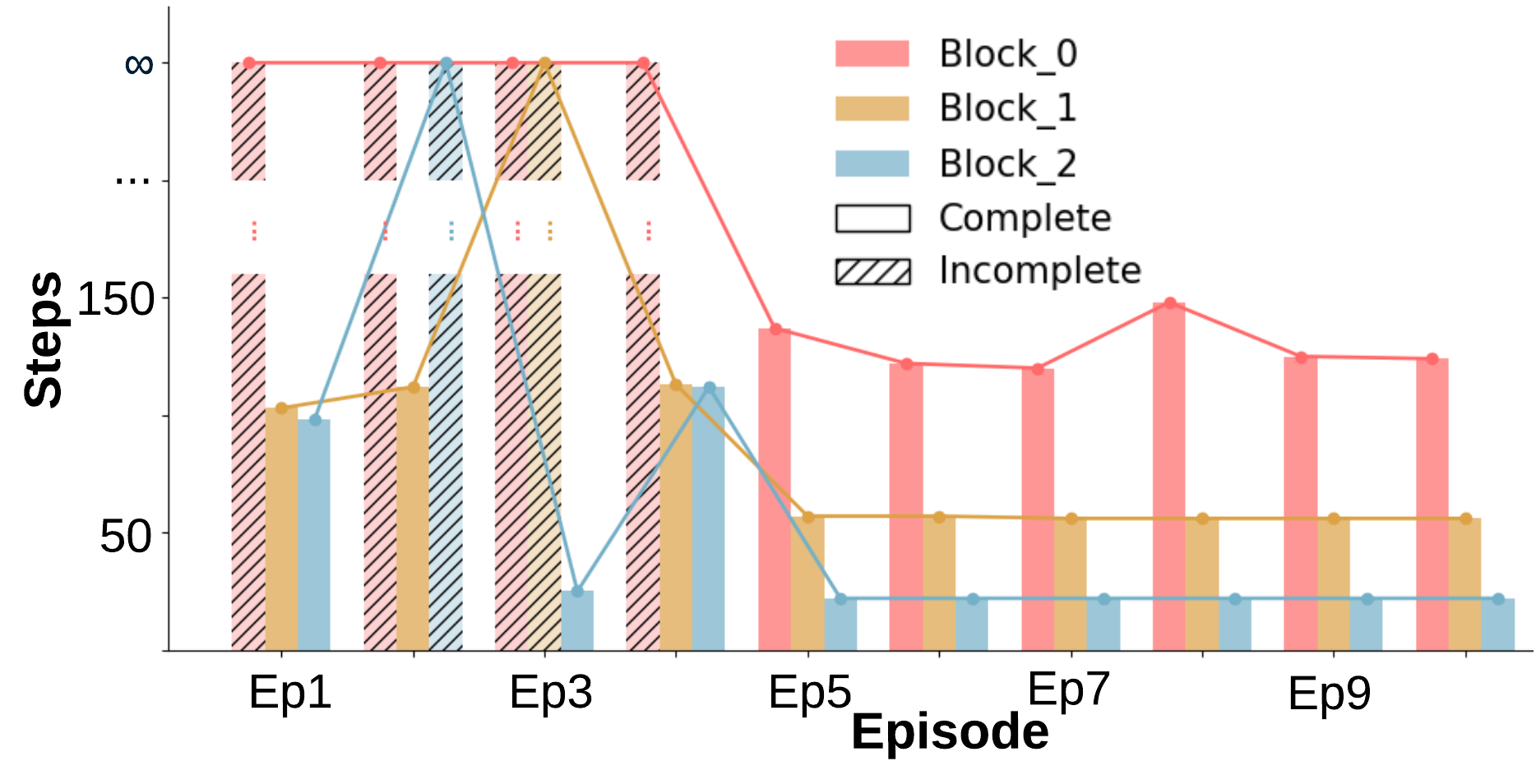}
    \caption{Completion time (Env-steps).}
    \label{fig:block_envsteps}
  \end{subfigure}
  \caption{Performance of DR.WELL agent:
(a) Block-level completion map shows that, unlike the baseline agent, nearly all blocks are completed consistently after the early episodes. (b) Completion times in wall-clock seconds and (c) environment steps display smoother and downward trends compared to the baseline, reflecting faster, more reliable block completion. }
  \label{fig:dr_results}
\end{figure}

\begin{wrapfigure}{r}{0.45\textwidth}
    \centering
\includegraphics[width=\linewidth]{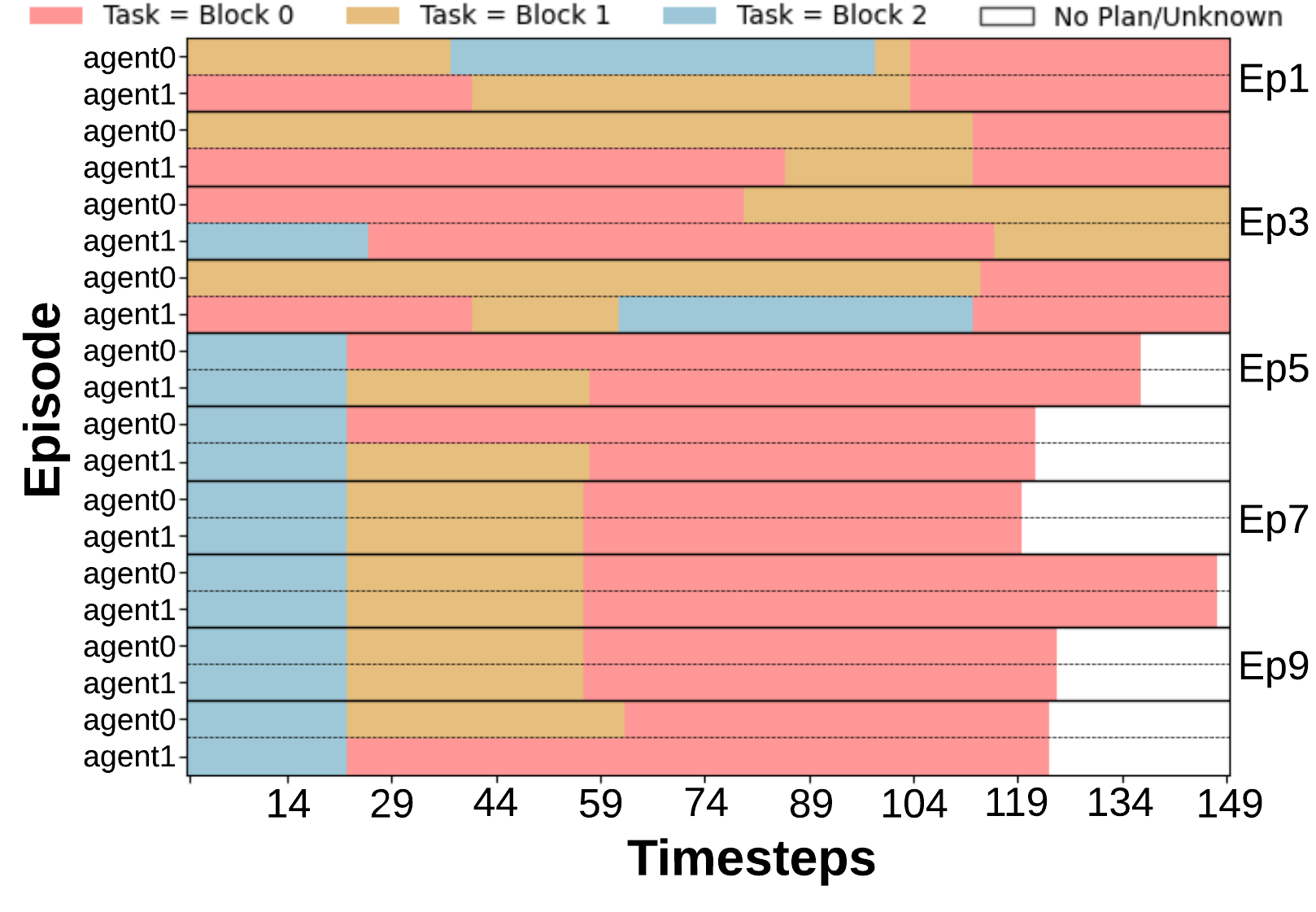}
\caption{Performance of DR.WELL: Task commitment patterns reveal how agents converge to stable allocations after Episode $5$, with less overlap and better division of labor.}
\label{fig:commit}
\vspace{-2mm}
\end{wrapfigure}

Figures~\ref{fig:dr_results} and~\ref{fig:commit} shows the results with our framework.  
Agents adapt their strategies across episodes by maintaining a shared symbolic world model and coordinating through communication and planning.
The block completion outcome heatmap (Figure~\ref{fig:block_completion}) shows that, 
unlike the baseline, almost all blocks are completed consistently after the early episodes. The completion timing trends (Figures~\ref{fig:block_timing} and \ref{fig:block_envsteps}) are smoother and, importantly,  show a clear downward trajectory, indicating that agents progressively adopt more effective strategies and achieve faster, more reliable block completion, especially for heavier blocks requiring multiple agents.

The task commitment patterns in Figure~\ref{fig:commit} reinforce this observation, showing that after Episode~$5$, agents converge on stable allocations with minimal overlap and improved division of labor.
 Although the wall-clock time increases slightly due to negotiation and replanning overhead, the overall number of environment steps declines, reflecting more deliberate coordination and increasingly efficient execution driven by the integration of our dynamic symbolic world model, plan revision, and communication.

Together, these results illustrate that maintaining a shared symbolic world model, combined with communication and planning, allows agents to adapt their strategies, enhance coordination, and achieve increasingly efficient outcomes over time.

\section{Conclusion and Future Work}\label{sec:conclusion}
We presented DR. WELL, a decentralized neurosymbolic framework for cooperative embodied multi-agent systems. DR. WELL is built around a structured two-phase negotiation protocol that enables agents to reach consensus under limited communication, and a dynamic world model (WM) that records commitments, plans, and execution traces into a symbolic memory. The symbolic representation functions as a structured space, in contrast to a latent space, where raw trajectories are clustered and compared to expose recurring patterns across episodes.
This design allows agents to coordinate without revealing full plans while grounding their reasoning in a shared representation.
Our experiments confirm that this combination of negotiation and symbolic memory yields cooperative behavior that is both reliable and interpretable. Compared to heuristic or naive LLM baselines, DR. WELL achieves higher completion rates and scales more effectively with agent count.

Future directions include extending sub-goal reasoning to capture latent steps, adapting observations to partial local views, supporting interruption and re-negotiation when plans fail, and enabling in-group communication during sub-tasks. Another direction is to make communication and task allocation more realistic, allowing agents to form strategies dynamically. Agents could also be given the ability to interrupt or redirect one another, fostering more adaptive coordination. Finally, incorporating probabilistic outcomes into the symbolic model would enable reasoning under uncertainty, better reflecting real embodied environments.

\section*{Acknowledgments}
This work was supported in part by the National Science Foundation (NSF) under grants CNS-2533813 and CNS-2312761. 



\small
\bibliography{references}
\bibliographystyle{plainnat}

\newpage
\appendix
\section{Supplementary Material}

\subsection{DR.WELL Formalization} \label{formalization}

Having outlined the core intuition behind the framework, we now formalize its main components.
We begin by formalizing the negotiation process, which constitutes the first stage of coordination in DR.WELL. 

The dynamic world model \(G_k = (V_k, E_k)\) functions as a shared symbolic memory that records outcomes of plans. Agents learn to coordinate and adapt collectively via two tightly coupled phases, \emph{communication} and \emph{planning}, each comprising two sub-stages. Together, these four stages form a recurrent cycle of collective reasoning and continual improvement.

At every timestep \(t\), let \(\mathcal{A}\) be the global agent set and define the subset of agents ready to replan:
\[
\mathcal{A}_t^{r} = \{\, a_i \in \mathcal{A} \mid a_i \text{ is ready to replan at time } t \,\}, \qquad
\mathcal{A}_t^{r} \subseteq \mathcal{A}
\]
If \(\lvert \mathcal{A}_t^{r} \rvert > 0\), these agents enter the \emph{communication room}. Communication proceeds in a \emph{round-robin} order defined by agent identifiers, \(\sigma_t = (a_1, \ldots, a_m)\).

\subsubsection{Negotiation.}

Negotiation operates in two sequential rounds: \emph{proposal} and \emph{commitment}. Both rounds are conducted through a shared communication buffer \(B_t^{(0)}\) that serves as a common workspace for all participating agents.

At the start of negotiation, the buffer is initialized with the current symbolic environment observation and a compact task summary retrieved once from the world model graph \(G_k = (V_k, E_k)\):
\[
B_t^{(0)} = \{\, s_t^{\text{env}},\ R_t = \{\,r(x) \mid x \in V_{\text{task}}^{\text{unique}}\}\,\}, \qquad
r(x) = [\,\hat{\mu}_{x}^{\text{start}},\ \hat{p}_{x}^{\text{succ}},\ \hat{\mu}_{x}^{\text{duration}},\ N_{x}^{\text{attempt}}\,]
\]
Here, \(r(x)\) contains empirical statistics accumulated across episodes, where \(\hat{\mu}_{x}^{\text{start}}\), \(\hat{p}_{x}^{\text{succ}}\), \(\hat{\mu}_{x}^{\text{dur}}\), and \(N_{x}^{\text{att}}\) denote the estimated mean start time, empirical success rate, mean duration, and total number of recorded attempts for task \(x\), respectively. These statistics are derived from plan instances grounded to task execution traces: each lowest-level node in \(G_k\) records completion outcomes and timing information, which propagate upward and are aggregated across episodes for all occurrences of the same task. The buffer \(B_t^{(0)}\) thus provides all agents with a shared cross-episode summary of task performance.

\textbf{(i) Proposal Round.}
Each agent \(a_j \in \sigma_t\) generates a proposal \(p_{a_j} \in V_{\text{task}}\) that includes a candidate task and its associated rationale, based on its current local information and the shared buffer:
\[
\forall j \in \{1, \ldots, m\}, \quad
p_{a_j} =
f_{\text{propose},a_j}\!\left(\phi_{a_j,t}, B_t^{(j-1)}\right),
\qquad
B_t^{(j)} = B_t^{(j-1)} \cup \{(a_j, p_{a_j})\}
\]
This round proceeds sequentially in the order \(\sigma_t = (a_1, \ldots, a_m)\), producing a distributed set of candidate task proposals that reflect the agents’ individual reasoning based on shared context.

After the proposal round, let \(X_t\) denote the set of unique tasks proposed. Using the current world model \(G_k\), the system retrieves, for each \(x \in X_t\), the empirical estimate of its optimal team size and the corresponding success rate:
\[
S_t = \{\,(\hat{n}(x),\, \hat{p}_{\text{succ}}(\hat{n}(x) \mid x)) \mid x \in X_t\,\},
\quad
\text{where} \quad
\hat{n}(x) = \arg\max_{n}\, \hat{p}_{\text{succ}}(n \mid x)
\]
with \(\hat{p}_{\text{succ}}(n \mid x)\) denoting the empirical probability that task \(x\) succeeds when executed by a team of size \(n\), as estimated from the current world model \(G_k\).

\textbf{(ii) Commitment Round.}
Each agent \(a_j \in \sigma_t\) determines its commitment \(c_{a_j} \in V_{\text{task}}\) based on its local information, the shared buffer, and the retrieved statistics. \(B_t^{(m)}\) is the buffer after the Proposal Round.
\[
\forall j \in \{1,\ldots,m\}, \quad
c_{a_j} =
f_{\text{commit},a_j}\!\left(\phi_{a_j,t}, B_t^{(m+j-1)}, S_t\right),
\qquad
B_t^{(m+j)} = B_t^{(m+j-1)} \cup \{(a_j, c_{a_j})\}
\]

Collectively, these individual decisions define the current agent-task mapping:
\[
M_t = \{(a_j, c_{a_j}) \mid a_j \in \mathcal{A}_t^{r}\}, \qquad M_t : \mathcal{A}_t^{r} \rightarrow V_{\text{task}}.
\]
The mapping \(M_t\) records the commitments and concludes the negotiation phase.

\subsubsection{Action Planning.}

Following negotiation, each committed agent independently constructs and refines an action plan for its assigned task. Unlike negotiation, this phase is fully decentralized: no further communication occurs among agents. Planning proceeds in two sequential stages: \emph{draft generation} and \emph{refinement}.

\textbf{(i) Draft Generation.}
Each agent \(a_j\) generates an initial plan prototype \(\pi_{a_j}^{\text{draft}}\) by reasoning over its local information and the current task mapping. The world model is not used at this stage (exploration proposal):
\[
\pi_{a_j}^{\text{draft}} =
f_{\text{draft},a_j}\!\left(\phi_{a_j,t}, M_t\right)
\]
The resulting draft corresponds to a plan prototype, which is an argument-free sequence of symbolic actions that defines the structural outline of the intended behavior.

\textbf{(ii) Refinement.}
Each agent then refines its draft plan using the world model \(G_k = (V_k, E_k)\), which stores aggregated planning outcomes across all episodes. For the agent’s committed task, the system first retrieves from \(G_k\) the top \(K\) historical plan prototypes ranked by success rate. For each retrieved prototype, it then retrieves the top \(L\) associated plan instances, ranked primarily by success rate and secondarily by shorter average duration.

\[
R_{x_{a_j}} = 
\{\,(\pi^{\text{proto}}, r^{\text{proto}}, 
\{(\pi^{\text{inst}}, r^{\text{inst}})|\operatorname{top}_{L}^{\text{inst}}\}) 
\mid \operatorname{top}_{K}^{\text{proto}}\,\},
\qquad
\pi_{a_j}^{\text{inst}} =
f_{\text{refine},a_j}\!\left(\pi_{a_j}^{\text{draft}}, R_{x_{a_j}}, \phi_{a_j,t}\right)
\]
Here, \(r^{\text{proto}}\) and \(r^{\text{inst}}\) denote retrieved summary statistics 
(e.g., success rate, mean duration, team size, or attempt count) aggregated at the prototype and instance levels, respectively. 
These statistics provide empirical guidance during plan refinement, allowing agents to balance exploitation of prior successful strategies with exploration of new action sequences.


\subsection{World Model Trace Information} \label{WM}

The world model records a full trace of each episode, and we include the complete log it produces in Section 2.3.
\lstset{
  basicstyle=\ttfamily\small,     
  numbers=left,                   
  numberstyle=\tiny\color{white},  
  stepnumber=1,                   
  numbersep=5pt,
  backgroundcolor=\color{gray!10},
  showspaces=false,
  showstringspaces=false,
  showtabs=false,
  frame=single,                   
  tabsize=2,
  breaklines=true,                
  breakatwhitespace=true
}
\begin{lstlisting}
============================================================
PHASE 1:COMMUNICATION
============================================================
=== PROPOSAL ROUND STARTED ===
=== CURRENT SESSION INFO ===
Current timestep: T15
Number of agents: 2

=== HISTORICAL TASK PERFORMANCE ===
Block_2: avg_start=T11.4, range=T0-T30, success_rate=54.5% (out of 11 attempts), min_completed_teamsize=1, min_task_duration=7.5, avg_task_duration=15.5

Block_1: avg_start=T29.4, range=T0-T50, success_rate=0.0% (out of 9 attempts), min_completed_teamsize=UNKNOWN, min_task_duration=UNKNOWN, avg_task_duration=UNKNOWN

Block_0: avg_start=T17.0, range=T0-T50, success_rate=0.0% (out of 5 attempts), min_completed_teamsize=UNKNOWN, min_task_duration=UNKNOWN, avg_task_duration=UNKNOWN

=== OPTIMAL TEAM SIZE RECOMMENDATIONS ===
Block_2: optimal team size = 1 (best success rate: 100.0%)
Block_1: optimal team size = UNKNOWN (no successful completions yet)
Block_0: optimal team size = UNKNOWN (no successful completions yet)

============================================================
PHASE 2: PROTOTYPES
============================================================
Planning for Block_0 (never completed)...

Historical Plan Prototypes (sorted by success rate):
1. Success rate: 0.0% | average team size 1.0 agents | unknown duration
Prototype: MoveToBlock -> Rendezvous -> Push

2. Success rate: 0.0% | average team size 1.5 agents | unknown duration
Prototype: MoveToBlock -> Rendezvous -> Push -> Rendezvous -> Push

3. Success rate: 0.0% | average team size 2.0 agents | unknown duration
Prototype: MoveToBlock -> Rendezvous -> Push -> YieldFace -> MoveToBlock -> Rendezvous -> Push

----------------------------------------

Planning for Block_1 (never completed)...

Historical Plan Prototypes (sorted by success rate):
1. Success rate: 0.0% | average team size 1.2 agents | average duration 19.7 steps
Prototype: MoveToBlock -> Push

2. Success rate: 0.0% | average team size 1.8 agents | average duration 18.0 steps
Prototype: MoveToBlock -> Rendezvous -> Push

----------------------------------------

Planning for Block_2 (successful completions)...

Historical Plan Prototypes (sorted by success rate):
1. Success rate: 60.0% | average team size 1.5 agents | average duration 14.2 steps
Prototype: MoveToBlock -> Rendezvous -> Push

2. Success rate: 0.0% | average team size 2.0 agents | average duration 15.0 steps
Prototype: MoveToBlock -> Rendezvous -> Push -> YieldFace

============================================================
PHASE 3: ENHANCED PLANNING - DETAILED INSTANCES
============================================================
Detailed instances for Block_0 (never completed)...

Detailed Plan Instances (sorted by success then duration):
1. Success rate: 0.0% | Attempts: 2 | Duration: unknown duration
Plan: MoveToBlock 0 left -> Rendezvous 0 left 2 10 -> Push 0 1
2. Success rate: 0.0% | Attempts: 1 | Duration: unknown duration
Plan: MoveToBlock 0 left -> Rendezvous 0 left 2 10 -> Push 0 1 -> YieldFace 0 left 1 -> MoveToBlock 0 left -> Rendezvous 0 left 2 10 -> Push 0 1

3. Success rate: 0.0% | Attempts: 1 | Duration: unknown duration
Plan: MoveToBlock 0 left -> Rendezvous 0 left 2 10 -> Push 0 1 -> Rendezvous 0 left 2 10 -> Push 0 1

4. Success rate: 0.0% | Attempts: 1 | Duration: unknown duration
Plan: MoveToBlock 0 left -> Rendezvous 0 left 2 5 -> Push 0 1 -> Rendezvous 0 left 2 5 -> Push 0 1

----------------------------------------

Detailed instances for Block_1 (never completed)...

Detailed Plan Instances (sorted by success then duration):
1. Success rate: 0.0% | Attempts: 1 | Duration: unknown duration
Plan: MoveToBlock 1 left -> Push 1 3

2. Success rate: 0.0% | Attempts: 1 | Duration: unknown duration
Plan: MoveToBlock 1 left -> Rendezvous 1 left 1 3 -> Push 1 1

3. Success rate: 0.0% | Attempts: 3 | Duration: 18.0 steps
Plan: MoveToBlock 1 left -> Rendezvous 1 left 1 5 -> Push 1 1

4. Success rate: 0.0% | Attempts: 4 | Duration: 19.7 steps
Plan: MoveToBlock 1 left -> Push 1 1

----------------------------------------

Detailed instances for Block_2 (successful completions)...

Detailed Plan Instances (sorted by success then duration):
1. Success rate: 100.0% | Attempts: 1 | Duration: 8.0 steps
Plan: MoveToBlock 2 left -> Rendezvous 2 left 1 3 -> Push 2 5

2. Success rate: 100.0% | Attempts: 2 | Duration: 9.0 steps
Plan: MoveToBlock 2 left -> Rendezvous 2 left 1 5 -> Push 2 7

3. Success rate: 100.0% | Attempts: 1 | Duration: 13.0 steps
Plan: MoveToBlock 2 left -> Rendezvous 2 left 1 5 -> Push 2 9

4. Success rate: 100.0% | Attempts: 1 | Duration: 22.0 steps
Plan: MoveToBlock 2 left -> Rendezvous 2 left 1 5 -> Push 2 11

5. Success rate: 100.0% | Attempts: 1 | Duration: 22.0 steps
Plan: MoveToBlock 2 left -> Rendezvous 2 left 1 10 -> Push 2 11

6. Success rate: 0.0% | Attempts: 4 | Duration: 14.8 steps
Plan: MoveToBlock 2 left -> Rendezvous 2 left 1 5 -> Push 2 1

7. Success rate: 0.0% | Attempts: 1 | Duration: 15.0 steps
Plan: MoveToBlock 2 left -> Rendezvous 2 left 1 5 -> Push 2 1 -> YieldFace 2 left 1
\end{lstlisting}

\pagebreak
\subsection{Dynamic Symbolic World Model} \label{WM_graph}

\begin{figure}[H]
    \centering
    \includegraphics[width=1\linewidth]{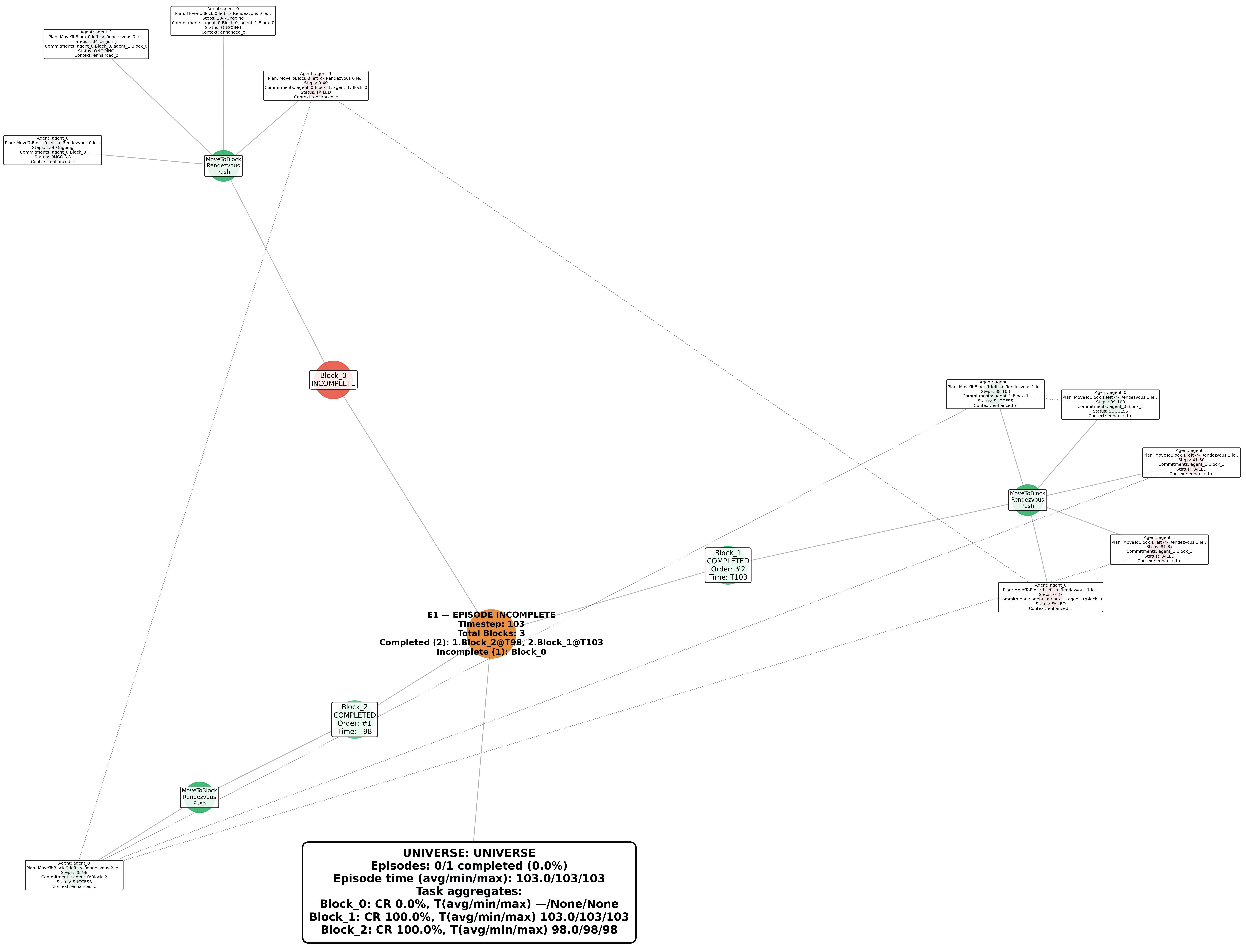}
    \caption{Full-resolution Dynamic Symbolic World Model (WM) snapshot at Episode 1. Nodes are color-coded by outcome: \textcolor{mgreen}{successful task/instance [green]}, \textcolor{mred}{block incomplete [red]}, \textcolor{morange}{episode incomplete [orange]} (i.e., at least one block undelivered). 
Edges indicate the relationships from the episode summary to task nodes, from tasks to their plan prototypes, and from prototypes to concrete plan instances. 
The episode summary node shows an incomplete outcome, with one block left unfinished. 
Only a small set of task nodes, plan prototypes, and instances are present, illustrating the sparse structure of the WM in early episodes. }
    \label{fig:E1}
\end{figure}

\begin{figure}[H]
    \centering
    \includegraphics[width=1\linewidth]{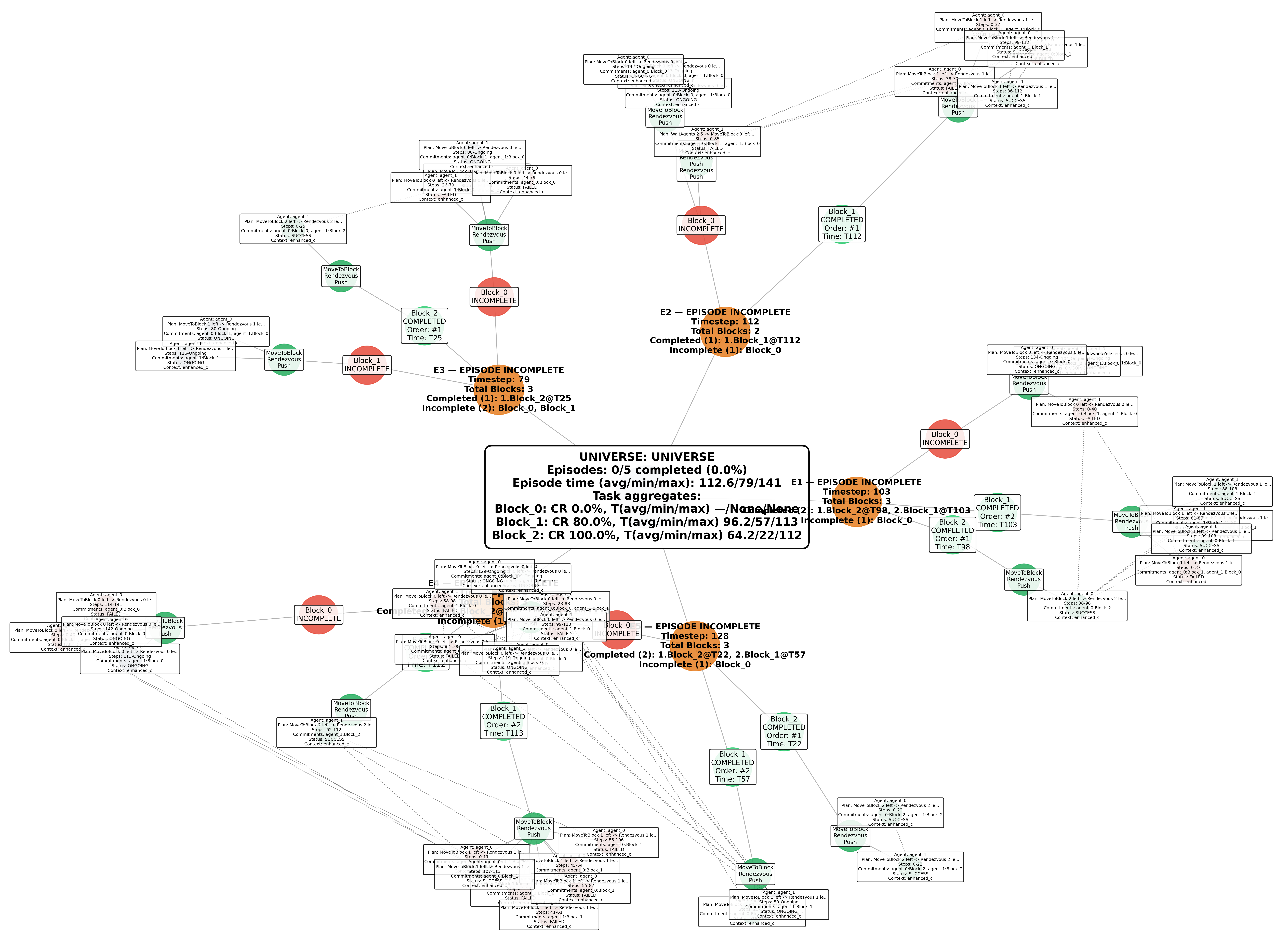}
    \caption{Full-resolution WM snapshot at Episode 5. Nodes are color-coded by outcome: \textcolor{mgreen}{successful task/instance [green]}, \textcolor{mred}{block incomplete [red]}, \textcolor{morange}{episode incomplete [orange]} (i.e., at least one block undelivered). 
Edges indicate the relationships from the episode summary to task nodes, from tasks to their plan prototypes, and from prototypes to concrete plan instances. 
The episode summary indicates partial completion, with agents recording multiple task attempts. 
The graph shows more plan prototypes and instances than in Episode 1, reflecting accumulated symbolic traces and alternative strategies. }
    \label{fig:E5}
\end{figure}

\begin{figure}[H]
    \centering
    \includegraphics[width=1.05\linewidth]{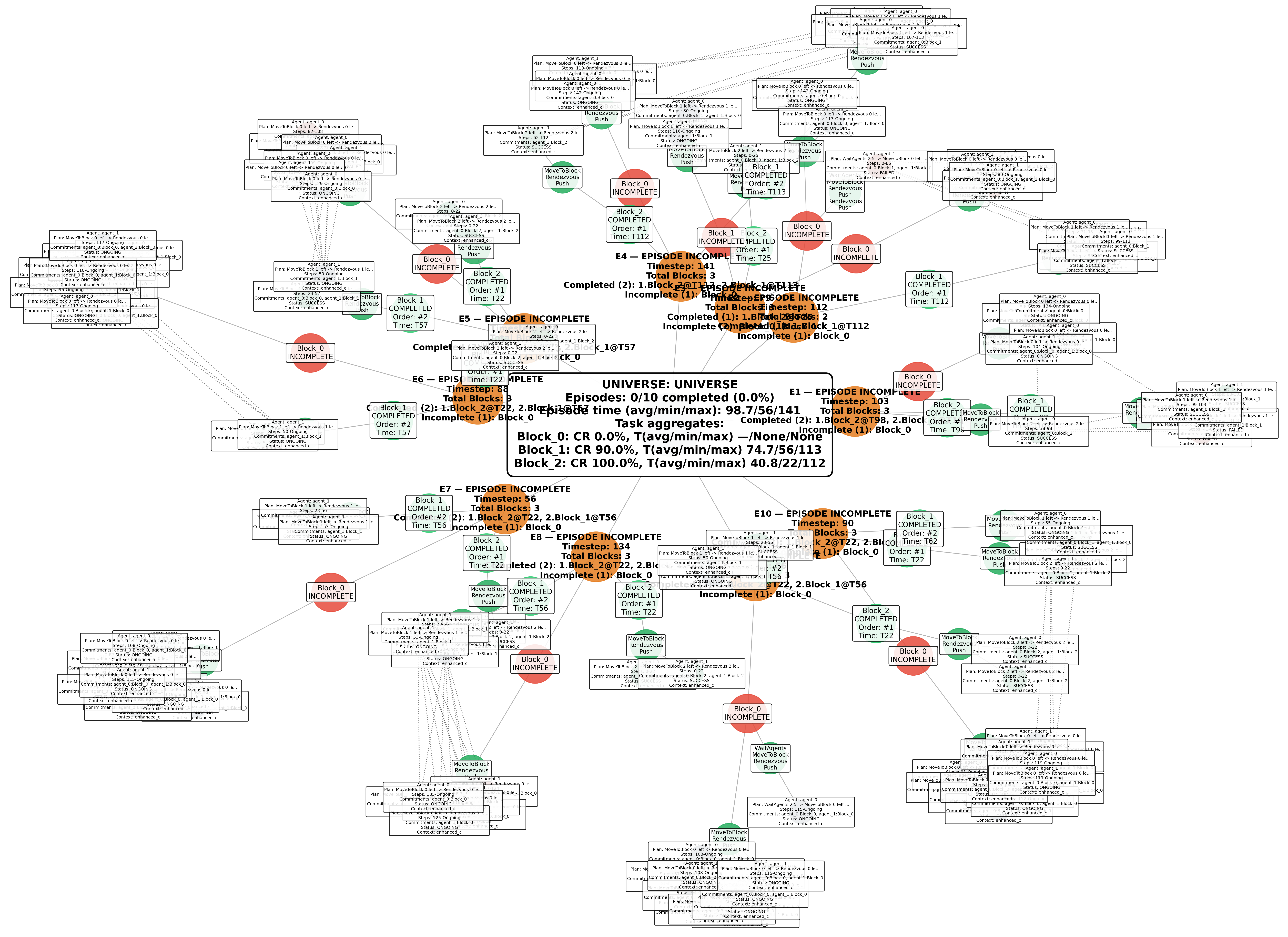}
    \caption{Full-resolution WM snapshot at Episode 10. Nodes are color-coded by outcome: \textcolor{mgreen}{successful task/instance [green]}, \textcolor{mred}{block incomplete [red]}, \textcolor{morange}{episode incomplete [orange]} (i.e., at least one block undelivered). 
Edges indicate the relationships from the episode summary to task nodes, from tasks to their plan prototypes, and from prototypes to concrete plan instances. 
The episode summary again reports incomplete delivery, but the graph is substantially larger. 
Tasks are linked to multiple prototypes and instances, capturing the expanded symbolic record of actions and outcomes by later episodes. }
    \label{fig:E10}
\end{figure}

\end{document}